\documentclass{article}

\usepackage{microtype}
\usepackage{graphicx}
\usepackage{booktabs} %
\usepackage{color}

\usepackage{hyperref}
\usepackage{float}
\usepackage{breqn}
\usepackage{booktabs}
\usepackage{enumitem}
\usepackage{caption}
\usepackage{subcaption}

\usepackage{amsmath,amsfonts,bm}

\def\eqref#1{equation~\ref{#1}}

\def\1{\bm{1}}

\def\rvx{{\mathbf{x}}}
\def\rvy{{\mathbf{y}}}

\DeclareMathAlphabet{\mathsfit}{\encodingdefault}{\sfdefault}{m}{sl}
\SetMathAlphabet{\mathsfit}{bold}{\encodingdefault}{\sfdefault}{bx}{n}

\definecolor{dgreen}{rgb}{0.0, 0.5, 0.0}
\definecolor{dblue}{rgb}{0.0, 0.53, 0.74}
\definecolor{dred}{rgb}{0.8, 0.0, 0.0}
\definecolor{dorange}{rgb}{1, 0.55, 0}

\newcommand{\ie}{\textit{i.e.}}
\newcommand{\eg}{\textit{e.g.}}

\newcommand{\versus}{\textit{vs.}}

\usepackage[accepted]{icml2024}

\usepackage{amsmath}
\usepackage{amssymb}
\usepackage{mathtools}
\usepackage{amsthm}

\usepackage[capitalize,noabbrev]{cleveref}

\theoremstyle{plain}

\theoremstyle{definition}

\theoremstyle{remark}

\begin{document}
\twocolumn[

\icmltitle{From Neurons to Neutrons: A Case Study in Interpretability}

\icmlsetsymbol{equal}{*}

\begin{icmlauthorlist}
\icmlauthor{Ouail Kitouni}{equal,iaifi,mit}
\icmlauthor{Niklas Nolte}{equal,meta}
\icmlauthor{Víctor Samuel Pérez-Díaz}{iaifi,harvard,cfa,urosario}
\icmlauthor{Sokratis Trifinopoulos}{iaifi,mit}
\icmlauthor{Mike Williams}{iaifi,mit}
\end{icmlauthorlist}

\icmlaffiliation{iaifi}{NSF Institute for Artificial Intelligence and Fundamental Interactions (IAIFI)}
\icmlaffiliation{meta}{FAIR at Meta}
\icmlaffiliation{mit}{Massachusetts Institute of Technology}
\icmlaffiliation{harvard}{Harvard John A. Paulson School of Engineering and Applied Sciences}
\icmlaffiliation{cfa}{Center for Astrophysics $\mid$ Harvard \& Smithsonian}
\icmlaffiliation{urosario}{School of Engineering, Science and Technology, Universidad del Rosario}

\icmlcorrespondingauthor{Ouail Kitouni}{kitouni@mit.edu}

\icmlkeywords{Machine Learning, ICML, Representation Learning, Mechanistic Interpretability}
\vskip 0.3in
]

\printAffiliationsAndNotice{\icmlEqualContribution} %

\begin{abstract}
Mechanistic Interpretability (MI) promises a path toward fully understanding how neural networks make their predictions. Prior work demonstrates that even when trained to perform simple arithmetic, models can implement a variety of algorithms (sometimes concurrently) depending on initialization and hyperparameters. Does this mean neuron-level interpretability techniques have limited applicability? 
We argue that high-dimensional neural networks can learn low-dimensional representations of their training data that are useful beyond simply making good predictions.
Such representations can be \textit{understood}  through the mechanistic interpretability lens and provide insights that are surprisingly faithful to human-derived domain knowledge. This indicates that such approaches to interpretability can be useful for deriving a new understanding of a problem from models trained to solve it. As a case study, we extract nuclear physics concepts by studying models trained to reproduce nuclear data.
\end{abstract}

\begin{figure*}
\centering
\begin{subfigure}{.5\linewidth}
    \includegraphics[width=\linewidth,trim=0 0 0 0,clip]{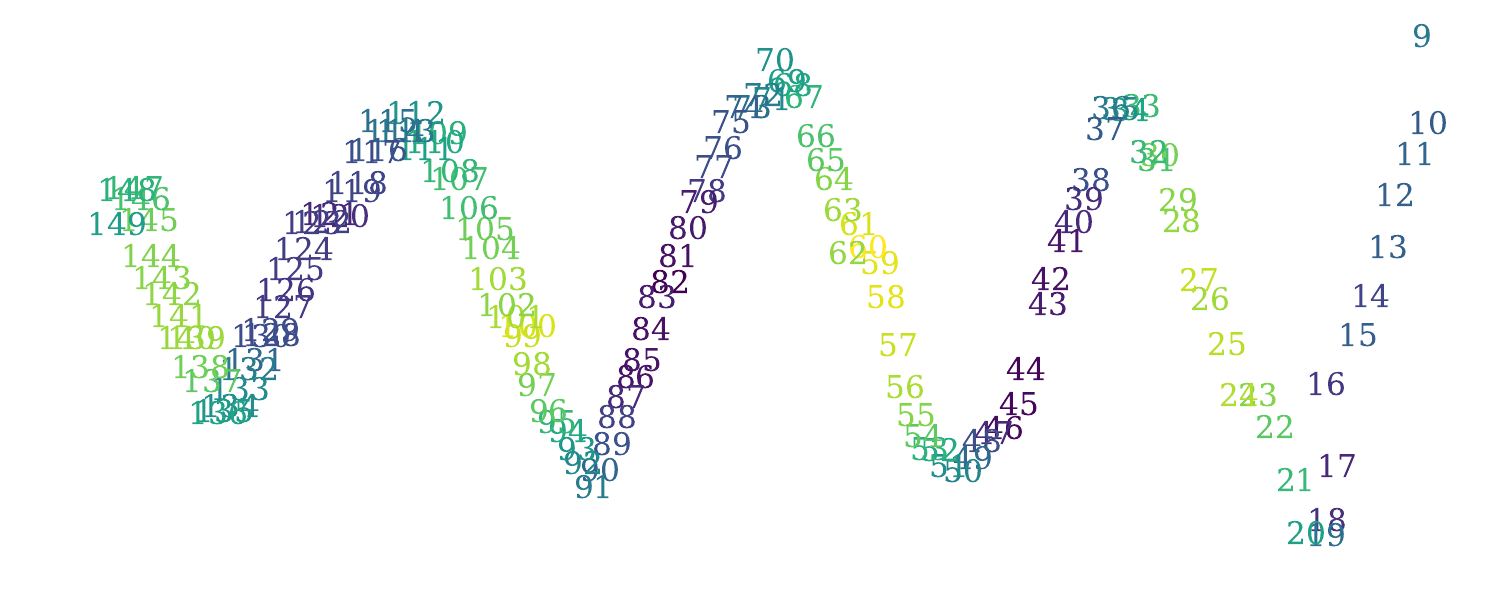}
\end{subfigure}%
\begin{subfigure}{.5\linewidth}
    \includegraphics[width=\linewidth,trim=0 0 0 0,clip]{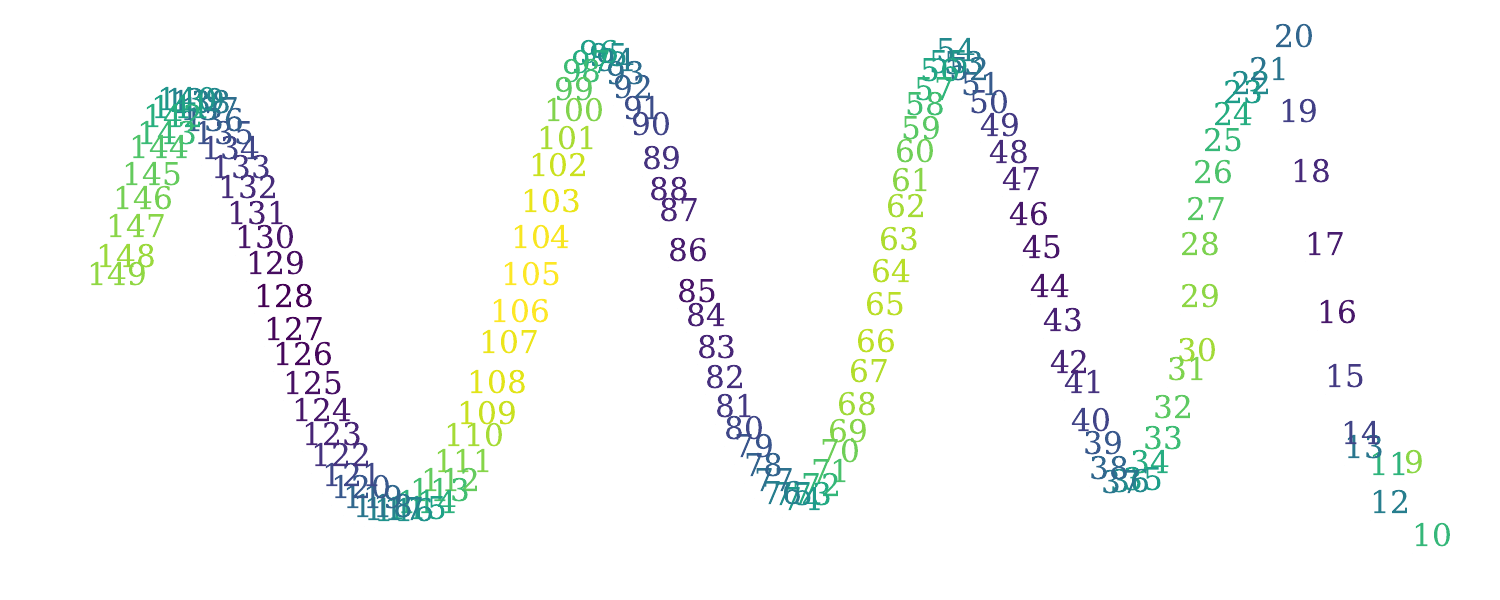}
\end{subfigure}
\caption{Projections of neutron number embeddings onto their first three principal components (PCs). Models were trained on nuclear data (left) or a human-derived nuclear theory (right). X-axis: 1st PC, Y-axis: 2nd PC, color: 3rd PC. Numbers indicate the neutron number ($N$) of each nucleus (see \textbf{Setup} in \cref{sec:beyond-arithmetic}). The helix structure encodes insights about nuclear physics discussed in subsequent sections.}
\label{fig:N-emb-binding}
\end{figure*}

\section{Introduction}
The scientific process involves understanding high-dimensional phenomena, often with large-scale data, and deriving low-dimensional theories that can accurately describe and predict the outcome of observations. There is mounting evidence that modern machine learning operates in a similar fashion, taking large-scale, high-dimensional data and deriving low-dimensional representations from them. For instance, recent work on the interpretability of deep learning has focused on understanding the low-dimensional representations learned by these models, with a particular emphasis on disentangled representations that separate the underlying factors of variation in the data \cite{bengio2013representation,higgins2018towards,locatello2019challenging}. Disentanglement aims to learn representations where each latent dimension corresponds to a semantically meaningful factor, such that varying one dimension while keeping others fixed produces interpretable changes in the input space \cite{burgess2018understanding,chen2018isolating,kim2018disentangling}.

Given the success of deep learning at modeling a wide variety of data, it seems plausible that  interpretability can help us learn from these models that are effectively domain experts.\footnote{There are of course some caveats here such as the question of the robustness of learned representations.} 
In this work, we investigate the ability of machine-learned algorithms to re-derive insights in human-developed understanding, taking nuclear theory as a case study of mechanistic interpretability.

Modern machine learning posits the manifold hypothesis~\cite{bengio2013representation}, the idea that most natural data we tend to care about lives in a low-dimensional manifold embedded in the high-dimensional measurement space. This is observed across modalities and, more recently, in language modeling where low-rank representations are ubiquitous in fully-trained large language models~\cite{hu2021lora, aghajanyan2021intrinsic, li2018measuring, 2023arXiv230514314D, 2023arXiv230310512Z}. Due to the nature of the data or the various implicit biases of the modern deep learning training procedures, neural networks learn compact representations that live in a small subspace of the inputs. Interpretability in deep learning has always been an active area of research \cite{kadir2001saliency,interp-survey} but the process of understanding how neural networks operate to make particular predictions (macroscopic phenomena) by uncovering the algorithms they implement (microscopic phenomena), is a nascent field of deep learning built around the idea that neural networks, despite their scale and complexity, can be interpreted and understood~\cite{elhage2021mathematical, olah2022mechinterp}. Here, we further posit that not only can they be understood, but they can also be used to say something useful about the nature of the problem they aim to solve. In the following, we will investigate whether mechanistic approaches can uncover scientific knowledge derived from the prediction task the model is trained on.
In other words, we propose expanding the view on MI from ``\textit{How does a model make predictions?}" to include ``\textit{What can the model tell us about the data?}"

In \cref{sec:modular-arithmetic}, we discuss prior work on MI in modular arithmetic and show an intuitive example of how it can be used to understand the algorithm that a simple MLP can learn to perform modular addition. Transitioning from modular arithmetic, \cref{sec:beyond-arithmetic} introduces the nuclear physics problem we will be tackling, explains the model architecture, and summarizes some key properties of the established physical models used by physicists.
Then, in \cref{sec:pca-interp} we motivate and explain the approach we take to interpret the models trained on the nuclear physics data. 
Finally, in \cref{sec:actual-mech-interp}, we interpret and extract ubiquitous concepts from the model representations and show that these are  similar to the most important human-derived concepts. For example, in \cref{fig:N-emb-binding} we show a spiral pattern that emerges in the model's representation when trained on nuclear data is similar to the one that arises when training instead on pseudo data obtained from a human-derived nuclear theory.

\section{Modular Arithmetic Primer}
\label{sec:modular-arithmetic}
A recent wave of research in interpretability has focused on algorithmic tasks such as arithmetic or checking the parity of a sequence. This has good reason: These datasets are extremely clean, arbitrary in size, and non-trivial enough to show a variety of interesting phenomena.
Models trained to perform modular arithmetic have been shown to yield relatively interpretable structures in their embeddings~\cite{liu2022towards}. Prior work has shown that the algorithms by which the trained models perform the task can be recovered precisely by understanding the model mechanistically at the activation and neuron level. Furthermore, this interpretation can be used to provide progress measures for the model's ability to generalize~\cite{nanda2023progress}.
Beyond these directions, we can leverage interpretability not only to understand models but also to extract knowledge from the training data. In this work, we explore this shift in perspective in a highly specialized domain.

\begin{figure}[b!]
\centering
\begin{subfigure}{.39\linewidth}
    \includegraphics[width=\linewidth,trim=500 0 25 30,clip]{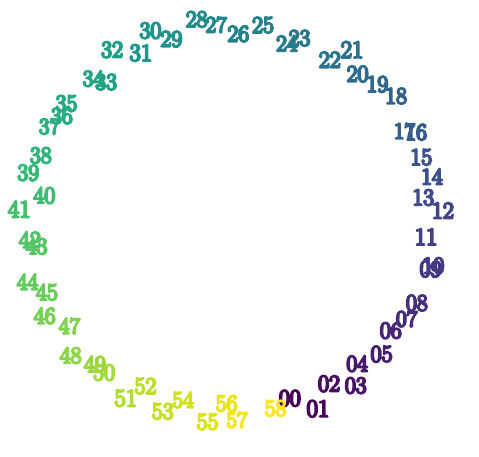}
\end{subfigure}%
\begin{subfigure}{.6\linewidth}
    \includegraphics[width=\linewidth,trim=0 0 0 0,clip]{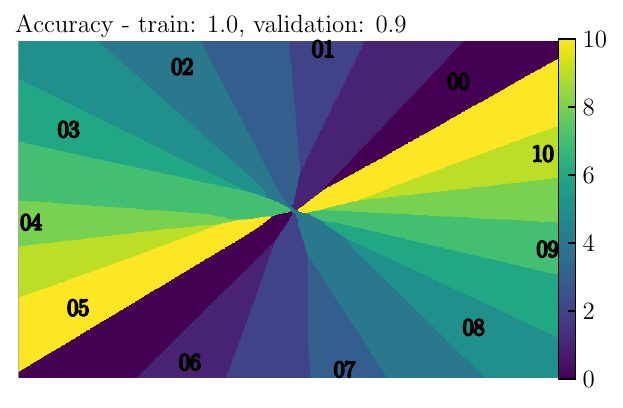}
\end{subfigure}
\vspace{-1em}
\caption{(left) Principal component projection of modular addition embeddings. The circular structure mirrors human-derived approaches used to teach modular arithmetic. (right) Model output in regions of the phase space. From \cite{liu2022towards}.}
\label{fig:pizza}
\end{figure}

First, we will revisit some of the mechanistic interpretability efforts for  models trained to perform modular addition. In \cref{fig:pizza} (left), we show the projection of the embeddings onto their first two principal components (PCs). 
Long after full generalization and circuit cleanup (see \citet{nanda2023progress} for a definition), the algorithm learned by the network involves a simple vector average. This can be visualized easily by projecting the first layer activations down to the first two principal components, uniformly sampling points in a two-dimensional grid, and feeding them back into the network after a reverse transformation to the right space. This procedure, which we will henceforth refer to as \textit{latent space topography} (LST), gives what the output of the network would have been as we move in a particular 2D subspace of the embeddings. As it turns out, this is quite informative. In \cref{fig:pizza} (right), we overlay the 2D projections of the embeddings for each integer on top of our latent space map and find that in order to compute the modular sum of numbers, the network first computes the vector average between the embeddings and returns the index of the slice the resulting sum falls into.
This fully explains the neural network solution to the problem but also sheds light on a new visual algorithm for modular addition. Simply arrange numbers around a circle, create slices between every two points, label the slices following the scheme given by the network in \cref{fig:pizza}, then finally obtain the sum of any two numbers by finding the mid point and reading off the label of the slice. 

In the following sections, we demonstrate the feasibility of knowledge extraction beyond modular arithmetic, using nuclear physics as a case study. Researchers have invested significant effort in understanding and modeling this domain over several decades. By training models on such data, we investigate whether known physics concepts can be identified through inspection of their representations.

\section{Beyond Arithmetic: A Physics Case Study}
\label{sec:beyond-arithmetic}

\paragraph{Why Nuclear Physics?} We choose to explore nuclear physics as a case study for several compelling reasons. First, physicists have studied various aspects of this data for decades and have developed simple yet effective expressions and concepts that explain the data well. This provides a useful frame of reference and a plausible approximate ``ground truth" for comparison. However, understanding the data remains a significant challenge, with several phenomena still unaccounted for by current theories and long-standing questions persisting. This combination of established knowledge and ongoing scientific challenges makes nuclear physics particularly interesting for interpretability research.
To further motivate our choice, consider a simple principal component projection in \cref{fig:N-emb-binding}, extracted the same way as \cref{fig:pizza} (left), but trained on nuclear physics.  A surprisingly periodic and continuous helical structure emerges, suggesting an opportunity for insightful interpretation.

The remainder of this section will be organized as follows:
First, we provide a description of the experimental process and the data to establish context. We also briefly discuss existing human-derived knowledge about the data.
Next, we take a close look at the input embeddings.  
Embeddings have been shown to carry significant structure in modular arithmetic training \cite{liu2022towards} and are a promising first step for model interpretation.
Finally, we study model features extracted from the penultimate layer activation and compare them to known physics terms to gauge similarities between model-derived and human-derived features.

\paragraph{Dataset and Nuclear Theory}

\begin{figure}[htbp]
    \centering
    \includegraphics[width=\linewidth,trim=0 0 0 0,clip]{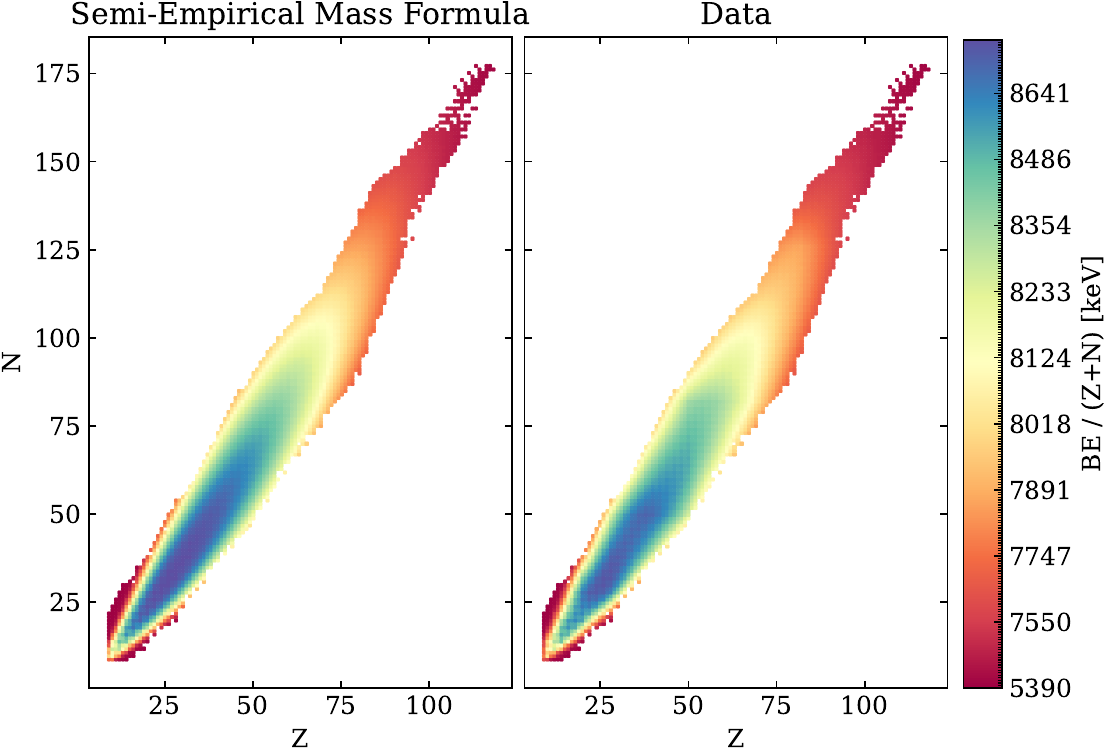}
    \caption{Binding energy per nucleon as given by the SEMF formula (left) and observed in measurements (right).}
    \label{fig:semf}
\end{figure}

Nuclei, the cores of atoms, have an array of interesting properties that depend on their composition. 
Like elements in the periodic table, they can be visualized on a two-dimensional grid and are characterized by two integer-valued inputs: the number of protons ($Z$) and neutrons ($N$), ranging from $1$ to $118$ and $0$ to $178$, respectively.
From these inputs, we aim to predict several continuous target properties of nuclei: binding energy ($E_{\rm B}$), charge radius ($R_{\rm ch}$), and various separation energies ($Q_{\text{A}}$, $Q_{\text{BM}}$, $Q_{\text{BMN}}$, $Q_{\text{EC}}$, $S_{\text{N}}$, $S_{\text{P}}$; see \cref{app:energies} for more details). As a form of regularization, we often also predict the input values $Z$ and $N$ that are obscured during embedding. This creates a multivariate regression task across up to $10$ target observables for $3363$ total nuclei. 
One of the most important nuclear observables is the binding energy. Many models have been developed in the literature with the liquid-drop model being the prototypical description of the nucleus. A consequence of the model is the renowned Semi-Empirical Mass Formula (SEMF) \cite{semf1935}:
\begin{align}\label{eq:semf}
   E_{\text{B}}=&\underbrace{a_{\text{V}}A}_{\rm Volume}
   -\underbrace{a_{\text{S}}A^{2/3}}_{\rm Surface}
   -\underbrace{a_{\text{C}}{\frac {(Z^2-Z)}{A^{1/3}}}}_{\rm Coulomb}\\\nonumber
   &\qquad\qquad  
   -\underbrace{a_{\text{A}}{\frac {(N-Z)^{2}}{A}}}_{\rm Asymmetry}
   +\underbrace{\delta (N,Z)~,}_{\rm Pairing}
\end{align}
where $A=N+Z$ is the total nucleon number. 
The coefficients $a_\ast$ are determined empirically. \cref{app:physics-models} contains more detailed explanations of each term. This formula is fairly accurate and theoretically well motivated. \cref{fig:semf} shows $E_{\text{B}}$ for both the data and the SEMF.

\paragraph{Setup}

We are interested in making predictions of the form $T(Z, N) = ?$, where $T$ is the task or observable being considered, and $Z$ and $N$ are integers uniquely identifying a nucleus on which predictions will be made.
Similar to the algorithmic tasks setup, inputs are tokenized and stacked in a sequence. Each token is embedded into a $d$-dimensional space. The sequence of embeddings $(E_Z, E_N, E_T)$ is then fed into the model which is tasked with completing the sequence using a numerical prediction. Specifically, the last token prediction is compared against the target numerical value and penalized with a mean-squared-error loss. Similar to~\citet{zhong2023clock}, we find that using attention provides a qualitatively different solution than input-independent attention~\citep{hassid2022much}. For the purposes of this paper, we will focus on fixed attention where all tokens are attended to equally\footnote{Without residual connections, this model could be written as a feedforward MLP.} (see \cref{app:training-model-details}). 

In all our experiments, we will consider one or several observables to predict with various models. The performance of the models will generally be measured by a Root-Mean-Square error (RMS) on a holdout set.\footnote{Error is in units of keV for energies and fm for lengths.} We will also predict some useful unitless quantities such as the neutron and proton numbers.

\paragraph{Objectives} Our goal will be to understand how the models' generalizing solutions work, extract useful representations from them, and compare those solutions to what is well-known in nuclear theory.
To ascertain the source of the learned representations, we can train our model on different tasks and collect results from the following experiments: \textbf{(1)} Train multiple models with different seeds on different data splits to understand the properties of generalizing versus memorizing solutions. \textbf{(2)} Study the internal representations of models trained on different tasks to understand the mechanistic effects of multi-tasking on generalization \ie\ what are the features of the representations that generalize and where do they come from?  \textbf{(3)} Compare the neural network-derived concepts with human-derived models.

\section{Are Principal Components Meaningful?}
\label{sec:pca-interp}
Principal Component Analysis (PCA) is a widely used dimensionality reduction technique due to its simplicity. However, it relies on several assumptions that, when violated, can result in erroneous conclusions. There is extensive literature discussing various PCA pitfalls, such as the complex relationship between oscillations and PCA~\citep{ novembre2008interpreting, antognini2018pca, lebedev2019analysis, proix2022interpreting}. Remarkably, these studies reported instances where non-oscillatory data exhibited oscillatory principal components. If this phenomenon is prevalent across various types of data, it is crucial to ensure it does not affect our results.

\subsection{Evidence 1: PCs Capture Most of the Performance}
\begin{figure}[htbp]
    \centering    \includegraphics[width=\linewidth]{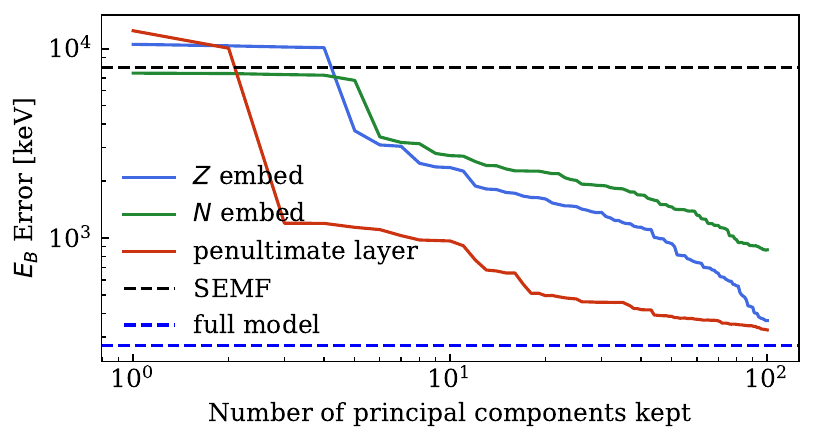}
    \caption{Binding energy prediction error as a function of number of PCs used at different layers.}
    \label{fig:pca_err}
\end{figure}

There is evidence in the literature that models operate on a much smaller subspace than their full dimension. Low-Rank adaptation \citep{hu2021lora} is an example showing that much of the performance gains from supervised fine-tuning can be obtained by training a low-rank approximation of the model.
If the PCs extracted were meaningless, we should see large performance gaps between the original model and one that solely relies on a subset of the PCs in making predictions.
However, we do indeed recover most of the performance with a relatively small number of PCs. \cref{fig:pca_err} shows the error as a function of principal components at different layers. To get this prediction, we project the activations (or the embeddings) onto their first $k$ principal components (ordered by variance) and set higher order components to zero. Then we invert the initial projection and consider the result the new activation that is sent through the rest of the network.

The behaviour observed in \cref{fig:pca_err} seems to be fairly universal, albeit to varying degrees. For instance, \citet{ashkboos2024slicegpt} recently utilized PCA to increase sparsity in language models by projecting activations to their principal components without losing significant performance.

\subsection{Evidence 2: Rich Structure}
\begin{figure}[htbp]
    \centering    \includegraphics[width=\linewidth]{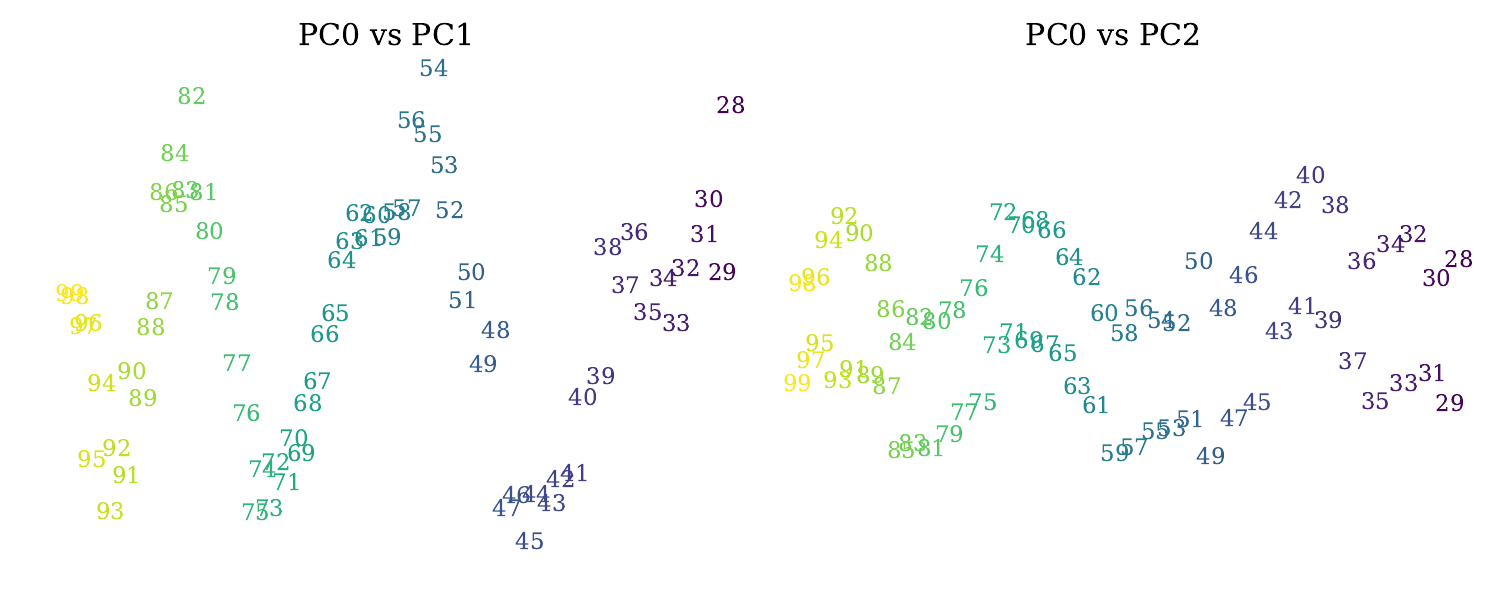}
    \caption{PC projections of Z embeddings from a model trained on all tasks. The color hue is a monotonic function of the proton number Z, to be able to quickly assess the presence of order.}
    \label{fig:rich-struct}
\end{figure}

\textit{Phantom oscillations} are sinusoidal patterns that can emerge in PCA even when the underlying data does not contain oscillations \citep{shinn2023phantom}. They can arise due to noise, smoothness across a continuum like time or space, or small misalignments/shifts across observations. 
Phantom oscillations characteristically emerge at multiple frequencies, with each principal component exhibiting a distinct frequency and lower frequencies explaining more variance.
In this work, we found that PC features exhibit unique patterns that differ from those expected in the case of noise. As observed in the previous section, highly informative structures emerge in the first two PCs of embeddings when learning modular arithmetic. Using \cref{fig:pizza} as a reference, \citet{liu2022towards} and \citet{zhong2023clock} hypothesized the complete algorithm used to perform the modular addition task. In the context of nuclear physics, similarly rich structures emerge during training beyond what would be expected in the case of noise. \cref{fig:rich-struct} displays the first two PCs of proton number embeddings extracted from a generalizing model. This clearly showcases features such as an even-odd split and periodicity, which we further explore in subsequent sections.

\section{Experiments}
\label{sec:actual-mech-interp}
\subsection{Embeddings}
Growing evidence, including studies on language model analogies(\eg, the ``$\textit{king} - \textit{man} + \textit{woman} = \textit{queen}$" analogy) \citep{mikolov2013efficient} 
suggests the presence of interpretable and robust structures in the initial embedding layers of neural networks. We can reasonably expect similar phenomena to occur in nuclear physics, and thus we will closely examine the neutron and proton number embeddings for trained models.

\begin{figure}[htbp]
    \centering
    \includegraphics[width=\linewidth,trim=10 10 10 10,clip]{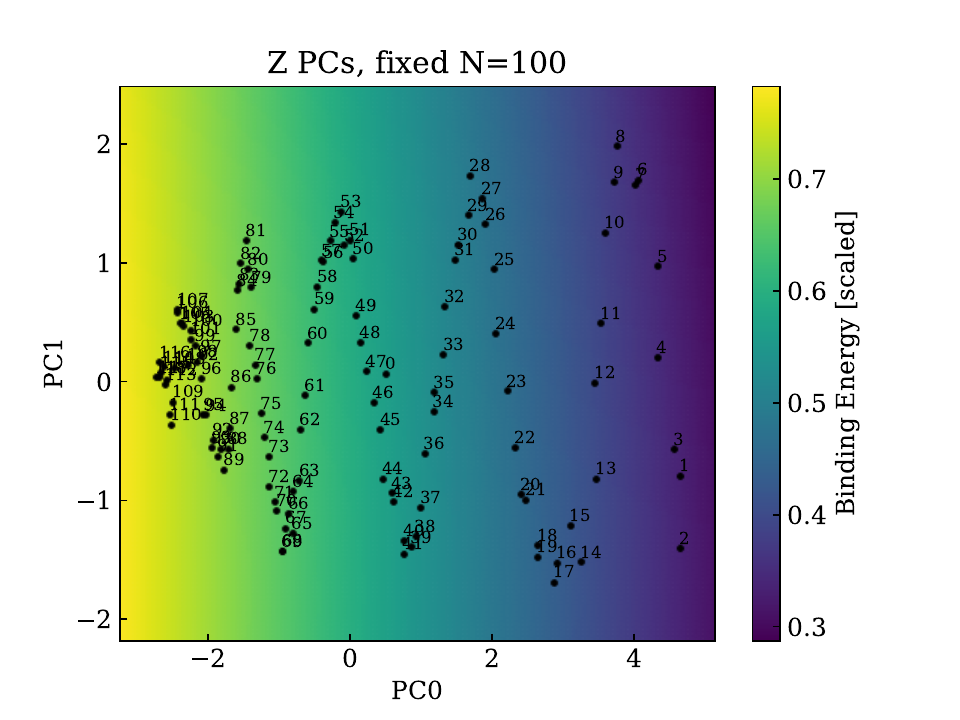}
    \caption{Projection of proton number ($Z$) embeddings onto the first two principal components (PCs), superimposed on the neural network's binding energy predictions. The binding energy LST is computed as a function of the first two PCs, while the remaining components are fixed at their mean values. Black dots indicate the positions of the $Z$ embeddings in this space, with the corresponding proton numbers annotated next to each dot. The color scale represents the predicted binding energy values, with brighter hues denoting higher energies.}
    \label{fig:response-pc01}
\end{figure}

Given the large dimensionality of the embeddings, we analyze the latent representations using a low-dimensional PCA projection, as motivated in \cref{sec:pca-interp}. 
\cref{fig:rich-struct} illustrates the three highest variance principal components of proton embeddings, plotted against each other. The observed structure, a helix (or spiral) pattern associated with increasing proton numbers, is one of the most striking features in the models trained. The color scheme transitions to lighter hues for higher numbers, emphasizing the clear numerical ordering observed.\footnote{While the number ordering could be expected for models where $N$ and $Z$ are among the prediction targets, it persists even in models where those targets are absent.} This ordering is also apparent, and the helix structure is particularly pronounced, in the high-variance primary components of the neutron number embeddings from \cref{fig:N-emb-binding}. Note that the color in this case represents the third PC.

\begin{figure}[htbp]
    \centering
    \includegraphics[width=.8\linewidth,trim=50 50 50 50,clip]{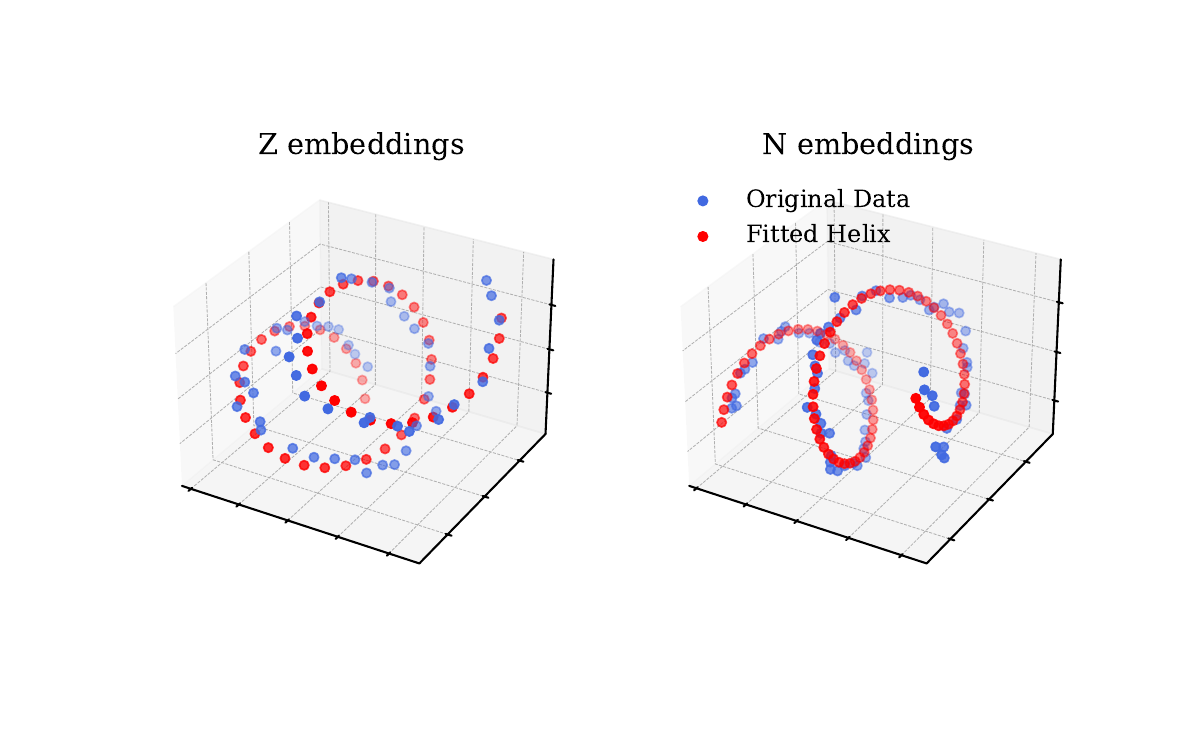}
    \caption{Fitting a helix to the PC-projected embeddings.}
    \label{fig:fit_spirals}
\end{figure}

Notably, $E_B$ has a strong correlation with both $N$ and $Z$, as seen in the first term of the SEMF. Therefore, it seems plausible that the inductive bias of ordering neutron and proton numbers in the embedding space is particularly beneficial.
To understand the model better, consider \cref{fig:response-pc01}, the latent space topography of $Z$ embeddings, constructed similarly to \cref{fig:pizza} for modular addition. It shows the predicted $E_{\rm B}$ as a colored background to the scatter plot of the two highest variance primary components in the $Z$ embeddings for $N=100$.
The dominating effect is the monotonic increase in binding energy when moving from right to left in PC0, which corresponds to the fact that $E_{\rm B}$ scales as $A=Z+N$ to leading order (this is known as the \textit{volume term }in the SEMF \cref{eq:semf}). 
\vspace{-0.3cm}
\paragraph{Properties of Models That Generalize Well}
Modifying the model architecture and hyperparameters significantly can result in different generalizing algorithms. We explore a small region of the algorithmic phase space and discover that generalizing solutions share a set of common properties, which we enumerate here.
\begin{figure*}[htbp]
    \centering
    \includegraphics[width=0.9\linewidth,trim=10 10 10 10,clip]{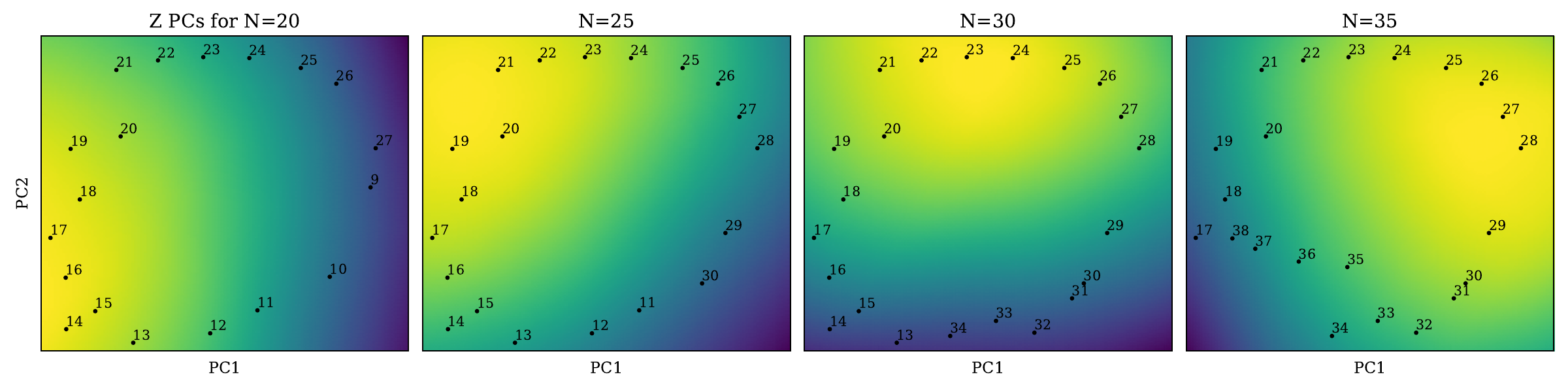}
    \caption{$Z$ embeddings projected onto principal components 1 and 2 (counting from 0) given multiple fixed neutron numbers. For each N, only Z embeddings are shown for which actual nuclei exist. The background shows the binding energy prediction of the model as a function of PC1 and PC2, where other primary components are fixed to their mean value. Brighter means more $E_B$.}
    \label{fig:response-pc12}
\end{figure*}
\vspace{-1em}
\paragraph{1. Helicity} 
We attempt to isolate the origin of the helix structure in the neutron and proton embeddings, and find that it represents a compelling geometric explanation of the data. Experiments reveal this structure appears when predicting binding energy. To elucidate how the model utilizes the helix, we parameterize it and perturb parameters to understand their effects (a detailed study with visualization is shown in \cref{app:spiral-mod}).
We fit a helix to the visually most helix-like portion of 3D PCA projections as illustrated in \cref{fig:fit_spirals}. The fits map to the projections well and enable us to isolate the effect of the different parameters of the helix.
For instance, we note that increasing the pitch (length of the central axis) elongates the helix, causing a constant offset in predictions, similar to the \textit{volume term} in the SEMF. Reducing the length has the opposite effect.
Increasing the radius ``sharpens" the downward arcs in predictions, likely linked to the SEMF's \textit{asymmetry term}, with radius controlling the prefactor.
The helix structure provides an interesting geometric explanation of how the model represents the data. In particular, it presents a complete description of the SEMF---itself motivated by geometry (\cref{app:LDM}) and basic physics principles---and yields particularly accurate fits, as shown in \cref{app:spiral-mod}.

\cref{fig:response-pc12} presents a complementary view to \cref{fig:response-pc01}, with the latent space topology displayed across the next two principal components (PC1 and PC2). This perspective is obtained by rotating the viewpoint by 90 degrees out-of-the-page compared to \cref{fig:response-pc01}.
For each pane, the neutron number ($N$) is fixed to a different value, increasing in increments of 5 between adjacent panes. The proton number ($Z$) embeddings displayed in each pane are limited to those corresponding to physically existing nuclei, \ie, $(Z, N)$ pairs present in the dataset.
The background is produced by evaluating the model by varying PC1 and PC2, keeping all other primary components fixed at their mean. We also tried varying PC0 but, as anticipated, we observed that changes in PC0, which aligns with the helix axis, only influence the absolute values of the model's output. 
The relative values within each LST ``slice'' remain stable.
Note that, since PC0 and $N$ are fixed, the overarching near-linear trend of binding energy with respect to increasing $N$ and $Z$ does not play a leading role here.

To focus on the local variations, we consider the binding energy relative to the nucleon number $A$ ($E_{\rm B}/A$) for the following analysis.
For each fixed $N$, there exists a specific $Z$ value that corresponds to the highest $E_{\rm B}/A$, representing the most stable element for that given $N$. As $Z$ diverges from this optimal value, the $E_{\rm B}/A$ decreases smoothly. This trend can be observed in Figure \ref{fig:semf}, where for each slice along the $N$ axis, there is a peak in $E_{\rm B}/A$ around a central $Z$ value (and \textit{vice versa} for slices along the $Z$ axis).
Consequently, for each $N$, there should be a continuous strip of $Z$ embeddings, with one embedding marking the highest $E_{\rm B}/A$ value, corresponding to the most stable nucleus for that particular $N$. Since each $N$ requires such a continuous strip, the entire sequence of $Z$ embeddings should form a continuous structure.

This is where the helix structure, which can be viewed as stacked circles, offers a compact and efficient way of achieving this continuity. By arranging the $Z$ embeddings along a helical path, the model ensures that for each $N$, there is a smooth progression of $Z$ values, with the most stable element located at the optimal position within the latent space. The helical structure allows for a continuous representation of the binding energy landscape, capturing the local variations and the stability peaks across different $N$ values.\footnote{See Appendix \ref{app:linear-continuity} for another example of continuity in the latent space.}

\begin{figure}[htbp]
    \centering
    \includegraphics[width=1\linewidth]{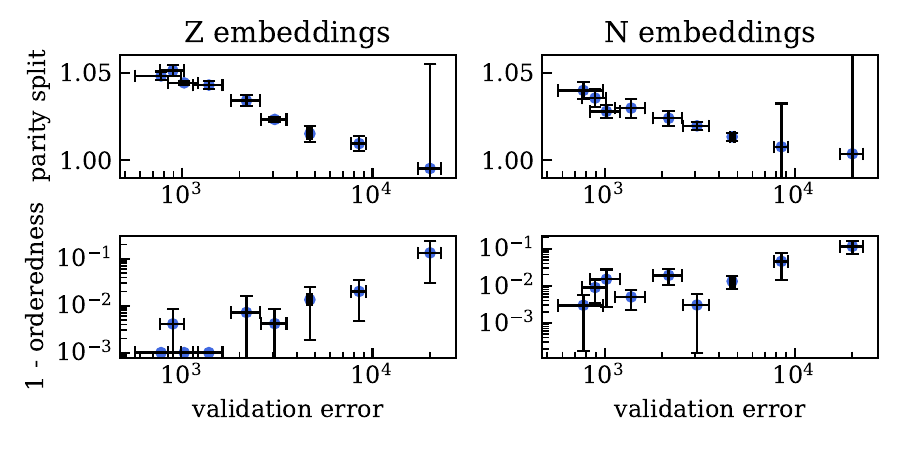}
    \caption{Parity split $R_{\text{P}}$ (top row) and $\text{orderedness}$ (bottom row) calculated on $N$ and $Z$ embeddings as a function of validation error. Zero values were clipped to $10^{-3}$ for visualization. Error bars are standard deviations and each point groups models trained with the same training fraction.}
    \label{fig:parity_orderedness_vs_perf}
\end{figure}

\paragraph{2. Orderedness}
We hypothesize that ordering numbers in the first few principal components is indicative of generalization and investigate the relationship between``orderedness" in embedding structures and generalization performance (see \cref{app:progress-measures} for the time evolution of this property). We train models with different train/validation splits (10\% to 90\% in 10\% increments, 3 random seeds each), varying batch size for consistent total optimization steps, and keeping other hyperparameters constant. 
Given the clear structure observed in the previous section, we experiment with a simple measurement of ordering along the first PC dimension. It reveals a surprising correlation with generalization performance, see \cref{fig:parity_orderedness_vs_perf}. 
We define the quantity,
$$
\text{orderedness} = \frac{1}{M}\sum_{i=1}^{M-1} \mathbf{1}(\tilde{\mathbf{E}}_0^i < \tilde{\mathbf{E}}_0^{i+1})~,
$$
where $\mathbf{1}$ is the indicator function,\footnote{The direction of the order might be reversed.} $\tilde{\textbf{E}}_0^i$ is the PC0 projection of the $N$ or $Z$ embedding, and $M$ is the total number of embeddings. We will generally use the tilde ($\,\tilde{\cdot}\,$) to denote PC-projected vectors.
It's important to note that all models fit the training data extremely well, with errors on the order of tens of keV. However, there is no correlation observed between train error and the degree of order.

\paragraph{3. Parity} 
In addition to orderedness, we explore another prominent feature in the embedding space: number parity. This feature is immediately apparent in the projection of PC0 and PC2 in \cref{fig:rich-struct} where even $Z$ embeddings are separated from odd $Z$ embeddings along PC2.
To measure the influence of parity on the embeddings, we introduce the following quantity:
\begin{equation*}
R_{\text{P}} = \frac{2\cdot d(\text{even},\text{odd})}{d(\text{even},\text{even}) + d(\text{odd},\text{odd})}~,
\end{equation*}
where $d(\cdot,\cdot)$ is the average pairwise $L_2$-distance between elements in the sets of even/odd $N$ or $Z$. This quantity is the ratio of the average distance of embeddings of different parity to that of embeddings of the same parity.
\cref{fig:parity_orderedness_vs_perf} illustrates how $R_{\text{P}}$, calculated on proton embeddings, correlates with validation performance.
The clear trend observed suggests that parity is an important indicator of model performance and possibly an important feature of the data.

\begin{figure}[htbp]
    \centering
    \includegraphics[width=\linewidth,trim=10 0 10 10,clip]{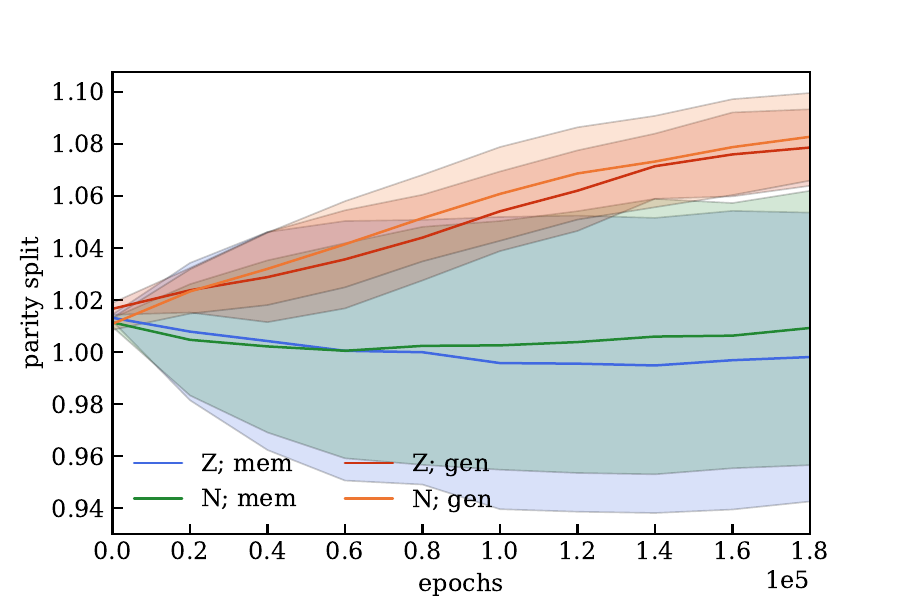}
    \caption{Parity split $R_{\text{P}}$ as a function of training time for $N$ and $Z$ embeddings for memorizing and generalizing models. The uncertainties are computed over 3 data and initialization seeds.
    }
    \label{fig:parity_vs_train_time} 
\end{figure}

It turns out that an important feature of nuclear properties is the tendency of nuclear constituents (both protons and neutrons) to form pairs.\footnote{This is related to the so-called Pauli Exclusion Principle \cite{Pauli1925}.} Numerous characteristics depend on the parity (even/odd) of $N$ and $Z$. This is evident in the \textit{Pairing} term of the SEMF, which changes sign based on the parity.

\subsection{Hidden Layer Features}
In the previous subsection, we explored proton and neutron embeddings to extract valuable information about models that generalize well. We discovered some properties of these models and were able to map them to well-known physics concepts. However, the functional relationship between initial embeddings and the output is often unclear. Now we focus on the activations of the penultimate layer, which does not have this drawback since it maps linearly to the output.
We continue to use PCA projections to visualize and analyze these high-dimensional features. As seen in \cref{fig:pca_err}, we can recover much of a model's performance using just a few of these features.
We observe that, similar to those we see in the embeddings, the principal components of the activations exhibit a rich structure, including terms that are smooth and slowly varying, others that have a high-frequency and small-scale, and some that are highly structured. Examples from each category are shown in the top row of \cref{fig:nice-features}, and a larger collection of PCs can be found in \cref{fig:last-layer-features} of the Appendix. 

We aim to recover human-derived descriptions of the problem in these latent representations, and we will do so based on a simple matching heuristic. 
Let $\tilde{\rvx}_i$ be the $i$-th vector of the neural network's penultimate layer features (given by the $i$-th PC dimension) and $\rvy_j$ be the $j$-th physical term vector produced by evaluating the term at all values of $N$ and $Z$ (see \cref{app:LDM,app:shell_model} for all terms). We use the cosine similarity, defined as $\mathrm{sim}(\tilde{\rvx}_i, \rvy_j) =  \tilde{\rvx}_i\cdot \rvy_j / ||\tilde{\rvx}_i|| || \rvy_j||$, to compare the two sets of vectors. We find that this heuristic recovers visually compelling matches and show a few examples in \cref{fig:nice-features} with the physical terms at the bottom and their matches in neural features at the top. We note the following:
\begin{figure}
    \centering
    \includegraphics[width=0.9\linewidth]{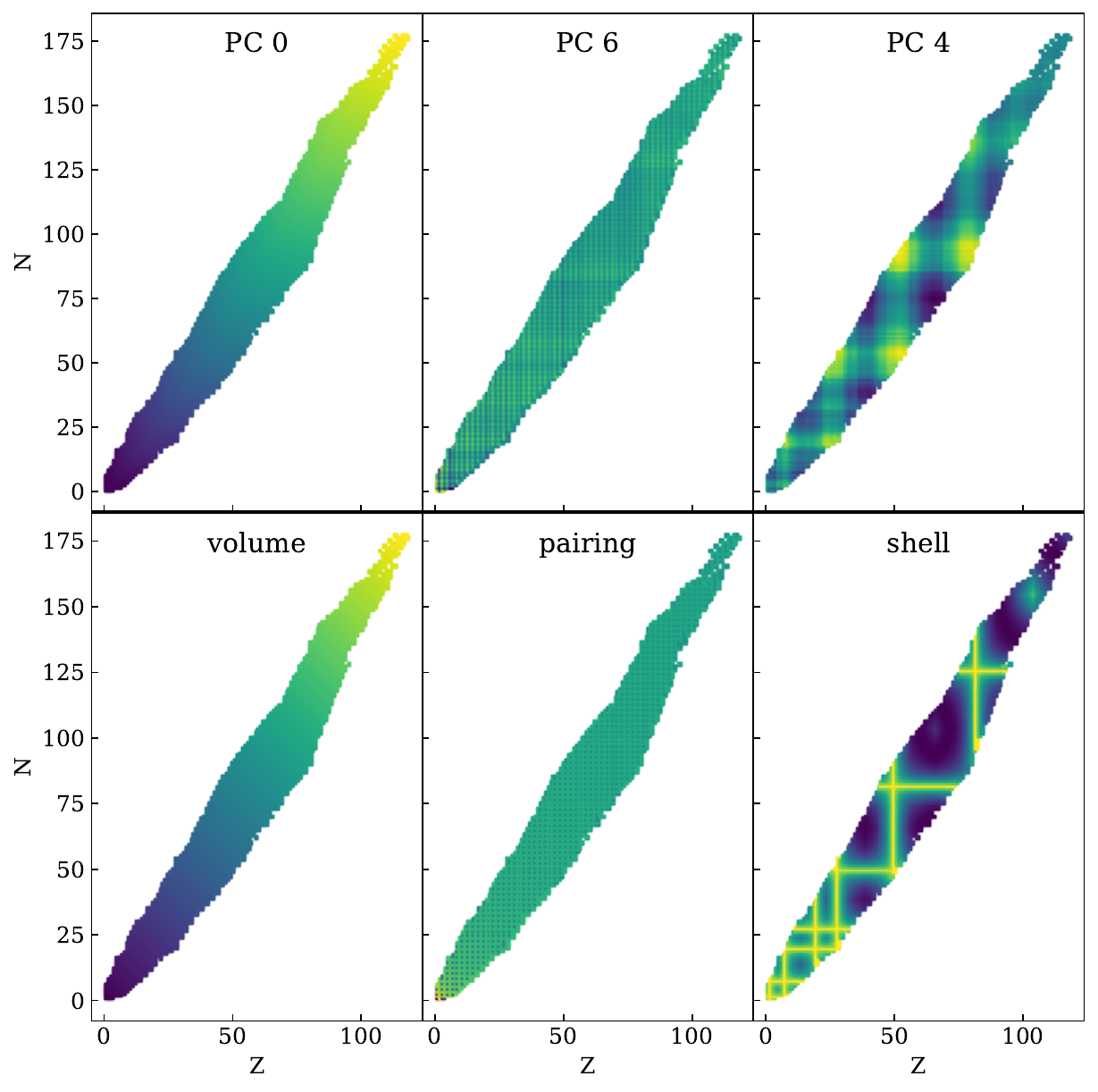}
    \caption{(Top) penultimate layer PCs and (bottom) physics terms with high similarity.}
    \label{fig:nice-features}
\end{figure}

\begin{itemize}[nosep,leftmargin=*]
    \item PC0 shows a strong trend towards higher values increasing $Z$ and $N$. Since the model predictions are linear combinations of those features, we can deduce that PC0 is primarily responsible for the general upwards trend in the output. Note the striking consistency of that trend with the effect of the PC0 of input embeddings (seen in \cref{fig:response-pc01}) and the number ordering described in the previous section.
    The bottom left pane of \cref{fig:nice-features} shows the dominant volume term of the SEMF, closely matching our feature PC0.
    \item Unlike PC0, the contribution of PC6 is of smaller scale, characterized by a high-frequency periodicity in both $N$ and $Z$.
Interestingly, we can also match this feature quite distinctly to the pairing term in the SEMF, observing that both are predominantly a function of the parity of $N$ and $Z$. Note again the close connection to the parity split observed in initial embeddings.
    \item Lastly, we take a look at PC4. This one stands out due to its obvious structure and the distinctive, staircase pattern.
    No term in the SEMF predicts this structure. As it turns out, a higher-order correction to the SEMF comes from the nuclear shell theory that predicts the significance of the so-called \textit{magic numbers} in $Z$ and $N$. The corresponding bottom-right pane  in \cref{fig:nice-features} shows the predicted contribution from the shell theory with strikingly similar structure as our PC4.
\end{itemize}
Note the significance of this finding: there is a vast amount of possible ways in which a neural network could decompose the problem, and yet, despite the simple techniques we used to inspect the activations, we were able to recover a range of human-derived concepts. With all of the above, we have (re)discovered the liquid drop model of nuclear physics and found hints of more advanced corrections from the shell model, simply by studying the weights and activations of a neural network trained on nuclear data.
We are currently working on further decoding what the machine has learned into human-interpretable knowledge.

\paragraph{Where Do These Representations Come From?}
\begin{figure}[htbp]
    \centering
    \includegraphics[width=\linewidth]{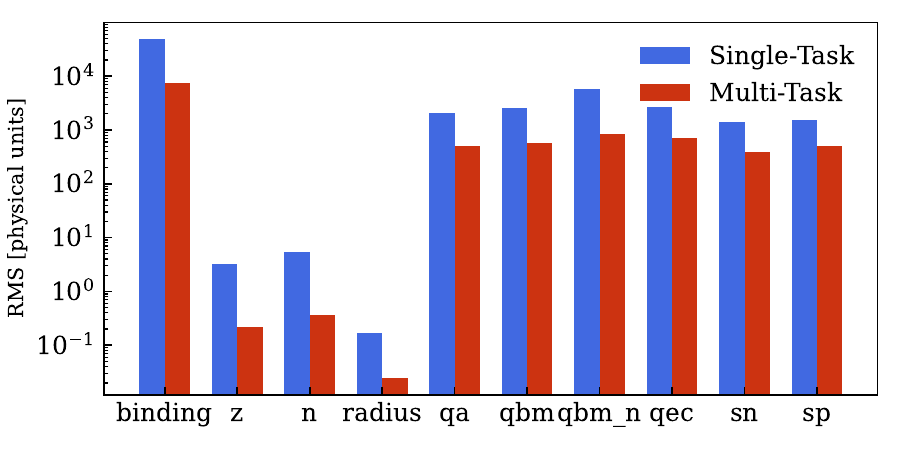}
    \caption{Test performance over different observables for models trained on a single task versus multiple tasks jointly.}
    \label{fig:MTLvsSTL}
\end{figure}
Learning from more diverse datasets should yield higher quality models and lead to improved generalization, provided that the model has enough capacity and nothing goes wrong with the training procedure. Naturally, this is expected to reflect also in the quality of the representations. \cref{fig:MTLvsSTL} demonstrates that using the same representations to predict a variety of nuclear observables improves the performance on each of them individually.
For this demonstration, we perform training runs with one feature at a time, or all at the same time, with 50\% of the data held out as a validation set in each setting to gauge the generalization performance. We observe a consistent improvement on all observables when tackling the problem with a multi-task solution, utilizing more data.

But where do the prominent features we observed in the latent representations come from? We systematically compare the representations learned on individual tasks and note that binding energy is primarily responsible for helicity and is never observed elsewhere, parity is most pronounced when training on separation energies, ordering seems to be partially present in many cases, and $Z$ and $N$ do not produce particularly interesting structures (examples in \cref{app:which-task-rep}). 

\paragraph{Symbolic Expressions for Discovering New Terms}
We can also use the latent representations to model what the neural network learned, and thus, extract a new physics model. We use symbolic regression to map to the features of the penultimate layer, and then apply a transformation that aligns to the binding energy. Using this pipeline we recover a predictive symbolic expression. The new formula achieves a better performance than the SEMF, though is less interpretable. As a baseline, we also regress directly over the task. However, we were not able to recover a performance as good as the one obtained exploiting the neural network features. Though in general, results would depend on the data, the model trained, and the symbolic regressor itself, this result suggests that the model learns to decompose the problem into features that can make it easier to find interpretable symbolic expressions. This is inline with prior work that derives symbolic formulae from neural network features for physical systems \citep{lemos2023rediscovering}.
See \cref{app:symbolic} for details.

\section{Related Work}
\label{sec:related}
As an emerging field, mechanistic interpretability has recently focused on large language models (LLMs) \cite{elhage2021mathematical}, but it is also starting to gain relevance in scientific discovery \cite{cranmer2023interpretable}. Another relevant line of work studies whether models build internal \textit{``world models''} \cite{2022arXiv221013382L, 2023arXiv231115930B, 2023arXiv230400612B}. Glimpses of more complex understanding have already emerged. For instance, LLMs have constructed (to some extent) knowledge in world geography~\cite{2023arXiv230600020R}, and meaningful representations of space and time~\cite{2023arXiv231002207G}, both of which have been studied since Word2Vec~\cite{mikolov2013efficient}. 

In computer vision, interpretability can take a more direct approach due to the visual nature of the data \cite{kadir2001saliency,simonyan2013deep}.
Here, mechanistic interpretability was used to gain insights on and improve the effectiveness of convolutional networks \cite{zeiler2014visualizing}. A more microscopic approach to layer level interpretability on vision models was explored in \citet{olah2017feature}.

\section{Conclusion}
\label{sec:conclusion}
In this work, we explore the potential of using mechanistic interpretability to extract scientific knowledge from neural networks trained on physics data. We not only investigate \textit{how} models make their predictions, but also \textit{what} insights the model can provide about the data. Our analysis has revealed several findings.
First, the learned embeddings of proton and neutron numbers exhibit interpretable structures such as the helix and parity splits, which are indicative of the models' generalization capabilities. These structures mirror known physics concepts like pairing effects, suggesting that the models are capable of learning and employing established scientific knowledge.
Second, our inspection of hidden layer activations has uncovered components that resemble terms in established theories: the semi-empirical mass formula and the nuclear shell model. 
This similarity in both macroscopic trends and microscopic structures suggests that the models are learning physically meaningful representations.
Finally, by employing latent space topography,\footnote{Example code is available here:\\ \url{https://github.com/samuelperezdi/nuclr-icml}}  we were able to arrive at a full description of the algorithms used by the model to make accurate binding energy predictions. In particular, we found that the learned embeddings provide a geometric representation of the theoretically well-motivated SEMF.
These findings provide a proof-of-concept that neural networks, when trained on scientific data, can learn useful representations that align with human knowledge. This opens up exciting possibilities for future research on richer data and more complex tasks, which may uncover new scientific insights.

\section*{Acknowledgements}
This work is supported by the National Science Foundation under Cooperative Agreement PHY-2019786 (The NSF AI Institute for Artificial Intelligence and Fundamental Interactions, http://iaifi.org/). ST is also supported by the Swiss National Science Foundation - project n. P500PT 203156. VSPD acknowledges support from NASA/Chandra AR3-24002X grant.

\section*{Impact Statement}

This section presents a brief overview of our vision for an MI-enhanced approach to the scientific endeavor. Throughout the history of science, natural laws have been discovered by domain scientists studying high-dimensional data and realizing that, in some cases, these data can be explained by a simple interpretable picture. These pictures were generated in the minds of the domain scientists, often based on a simplified geometrical model of the system being studied.

We present a new approach to generating interpretable models from scientific data: rather than having domain experts study the high-dimensional data directly, we propose to first determine if a low-rank structure can be found in a machine-learned model representation. If it can, human domain scientists can try and decode this structure into an interpretable model, rather than continuing to work directly with the high-dimensional data.

Here, we chose an example where a human-derived interpretable picture is known to exist---nuclear physics and its famous Shell Model---and find that representation learning (without any physics input), along with the use of PCA, does indeed discover a low-rank geometric structure. After further study, using the Shell Model as a known baseline solution, we see that the machine has learned the Shell Model---though with corrections that lead to more precise predictions than the Nobel Prize-winning human-discovered model. Therefore, the known interpretable human-discovered model is found by the machine and communicated to us, albeit in a different form that still needed decoding by domain experts.

As in the nuclear physics case studied here, most human-discovered interpretable scientific models are only approximately true. In such cases, our approach has the potential to derive corrections to the human-discovered model, represented as deviations in the low-rank structure. We see this with the nuclear data and are working on fully decoding these deviations into interpretable correction terms to the Shell Model.

Such interpretable corrections will have a huge impact on the field of nuclear physics. This is especially true for exotic nuclei far from the stability region, which are impossible to make and study in the lab. Yet, the properties of these nuclei are crucial for understanding nuclear processes in extreme environments, such as neutron stars. This understanding, in turn, enhances our knowledge of how heavy elements were produced in our universe. This is an out-of-distribution (OOD) problem from the ML perspective, hence finding interpretable corrections that can be trusted in the OOD region is crucial.

Most other known interpretable models (in other scientific domains) are also only approximate, and similar corrections could likely be found to improve scientific knowledge in those areas as well. Furthermore, in many scientific domains, humans have not been capable of developing any interpretable theories, even approximate ones, when studying high-dimensional data. Whether our approach could lead to discoveries in such fields is impossible to predict---interpretable models may not exist for some highly non-linear problems---but it is a direction worth pursuing. Hence, one of our goals is to encourage the ML community to work more closely with domain scientists on such problems, which can drive a disproportionate impact across disciplines.

In summary, our work underscores the value of interpretability in scientific exploration. By elucidating how models represent problems, interpretability becomes a powerful tool for scientific discovery. As we continue to develop and refine these techniques, we anticipate that they will play an increasingly important role in advancing human understanding in a wide range of domains.

\bibliography{bib}
\bibliographystyle{icml2024}

\newpage
\appendix
\onecolumn
\section{Why does the model learn a helix?}
\label{app:spiral-mod}
The helix structure observed in the embeddings of both neutron and proton embeddings presents one of the most striking features in the model trained on nuclear properties. In an effort to get to the bottom of it, we attempt to isolate where it comes from. From experiments in the multi-task \versus single-task settings, we notice that having the binding energy as a target is a strong predictor for the appearance of the helix. Therefore we will restrict ourselves to the prediction of binding energy. 
Our strategy for shedding light on how the model uses the helix structure to its advantage is parameterizing and then perturbing the helix parameters. We hope to be able to factorize contributions from different aspects to break the process into understandable pieces. We fit a helix with trainable parameters using the following parametric equation:
\begin{equation}
\vec{r}(t) = R \left[ \cos(2\pi f t + \phi) \vec{u} + \sin(2\pi f t + \phi) \vec{v} \right] + P \vec{a} t  + \vec{r}_0~,
\end{equation}
where $\vec{u}$ and $\vec{v}$ are orthonormal unit vectors perpendicular to the central axis pointing towards the direction given by the unit vector $\vec{a}$. The shape parameters are: the length of central axis $\vec{P}$, the frequency $f$, the phase $\phi$, the radius $R$, and the origin $\vec{r}_0$. The direction of the evolution is chosen to be towards the visually most helix-like portion of 3D PCA projections of both neutron and proton embeddings.

In an effort to maximize visual clarity, we show experiments for a model trained on binding energy predictions from the SEMF, where we find a cleaner helix structure than when training on real data, see \cref{fig:N-emb-binding} (right). We constrain ourselves to $N \in [40, 120], Z \in [25, 80]$ to be able to fit the helix with a constant radius. The results of the fit can be found in \cref{fig:fit_spirals_semf}. The fits match the PC projections well and we can now perturb helix parameters. For visualization, we provide three plots for each parameter change: First, a plot of the helix with and without the changed parameter.
Second, the model prediction relative to $A=N+Z$ with and without the changed parameter as a function of $N$ for a fixed value of $Z$. Third, the same plot with $N$ and $Z$ roles reversed. We find that plotting relative to $A$ gives visually more informative results.

First, we increase the length parameter in \cref{fig:spiral_mod_P_1.2}. This elongates the helix along its main direction. Similarly as depicted in \cref{fig:response-pc01}, we find that moving along the main direction corresponds to a macroscopic term akin to the volume term in the SEMF.
Since we plot relative to $A$, that term causes, in first order, a constant offset in the predictions.
\cref{fig:spiral_mod_P_0.8} shows a reduction of the length, resulting in a negative offset.

Next, we increase the radius parameter, see \cref{fig:spiral_mod_R_1.5}. This causes the downwards facing arcs to ``sharpen". Taking a closer look at the SEMF formula and the $N$ vs. model output plot, we hypothesize that the depicted arcs are in fact the approximate parabola described by the third term and that the radius controls the prefactor of that parabola, causing the ``sharpening", or, in case of a radius parameter reduction, the flattening depicted in \cref{fig:spiral_mod_R_0.5}.

Lastly, we double the frequency parameter, see \cref{fig:spiral_mod_F_2}. There is no clear correspondence to any one particular term in the SEMF, but it gives an indication about how the arc is created. Doubling the frequency doubles the frequency of a now periodic sequence of arcs. This can be understood intuitively when observing  \cref{fig:response-pc12}. The ring structure with double frequency goes around twice and two periods appear in the model output. \cref{fig:spiral_mod_F_3} shows that this trend is persistent also when increasing the frequency even more.

While we have made decent progress towards understanding how the embeddings map to the output of the model, the full picture is not completely clear yet. However, we are confident that an iterative approach can help us understand the story completely.

{\onecolumn{}

\vspace{-1em}

\begin{figure}
    \centering
    \begin{subfigure}{.49\linewidth}
        \centering
        \includegraphics[width=\linewidth,trim=100 0 50 0,clip]{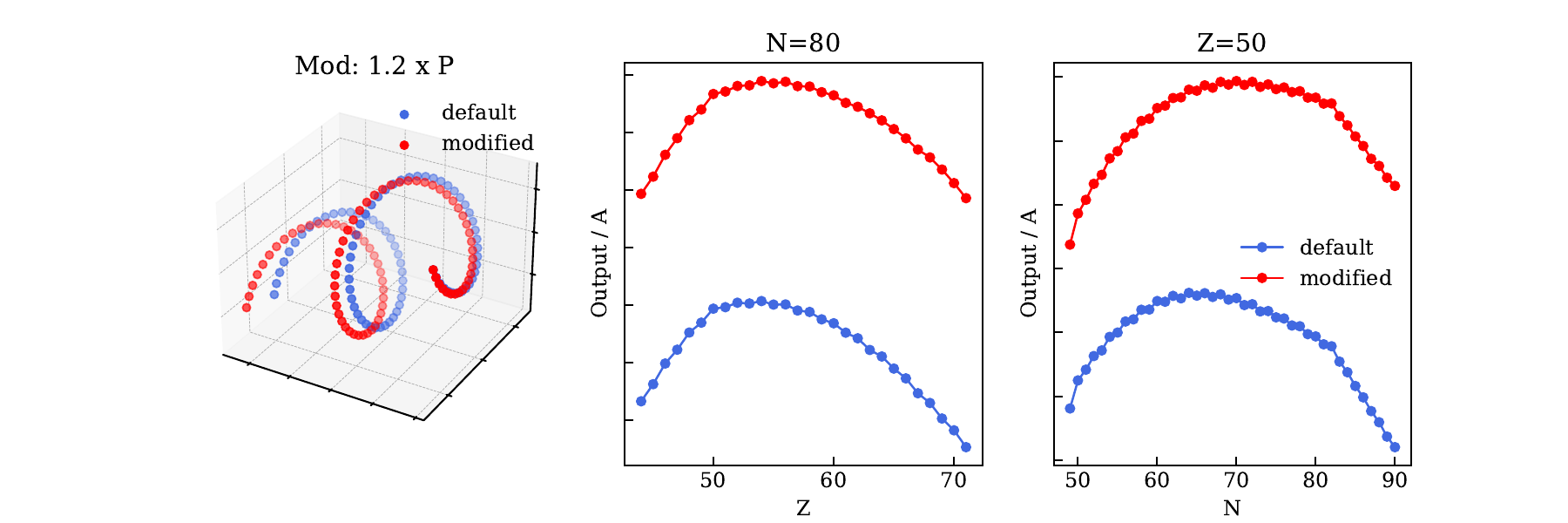}
        \caption{}
    \end{subfigure}
    \begin{subfigure}{.49\linewidth}
        \centering
        \includegraphics[width=\linewidth,trim=100 0 50 0,clip]{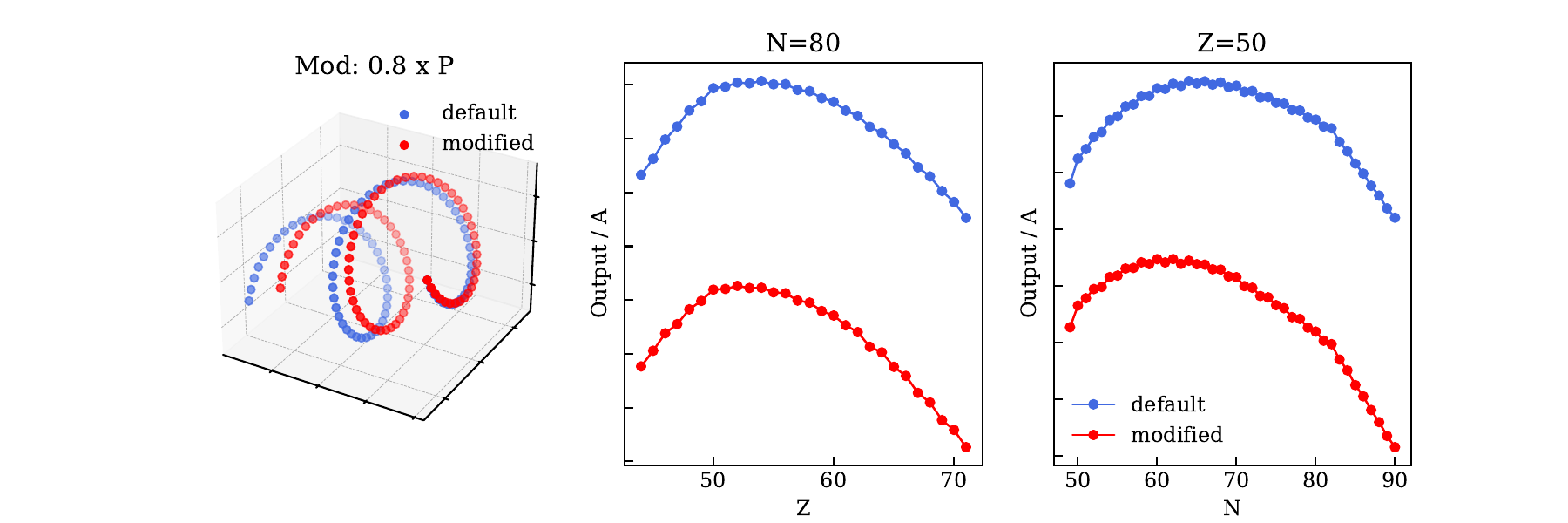}
        \caption{}
    \end{subfigure}
    \hfill
    \begin{subfigure}{.49\linewidth}
        \centering
        \includegraphics[width=\linewidth,trim=100 0 50 0,clip]{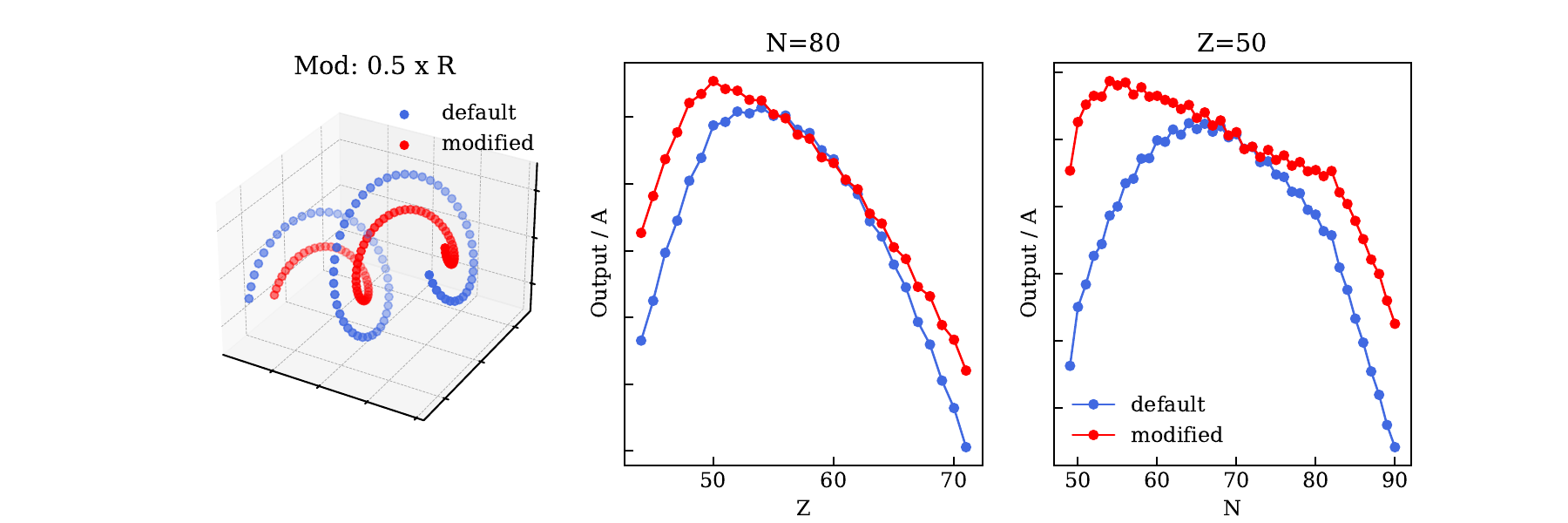}
        \caption{}
    \end{subfigure}
    \begin{subfigure}{.49\linewidth}
        \centering
        \includegraphics[width=\linewidth,trim=100 0 50 0,clip]{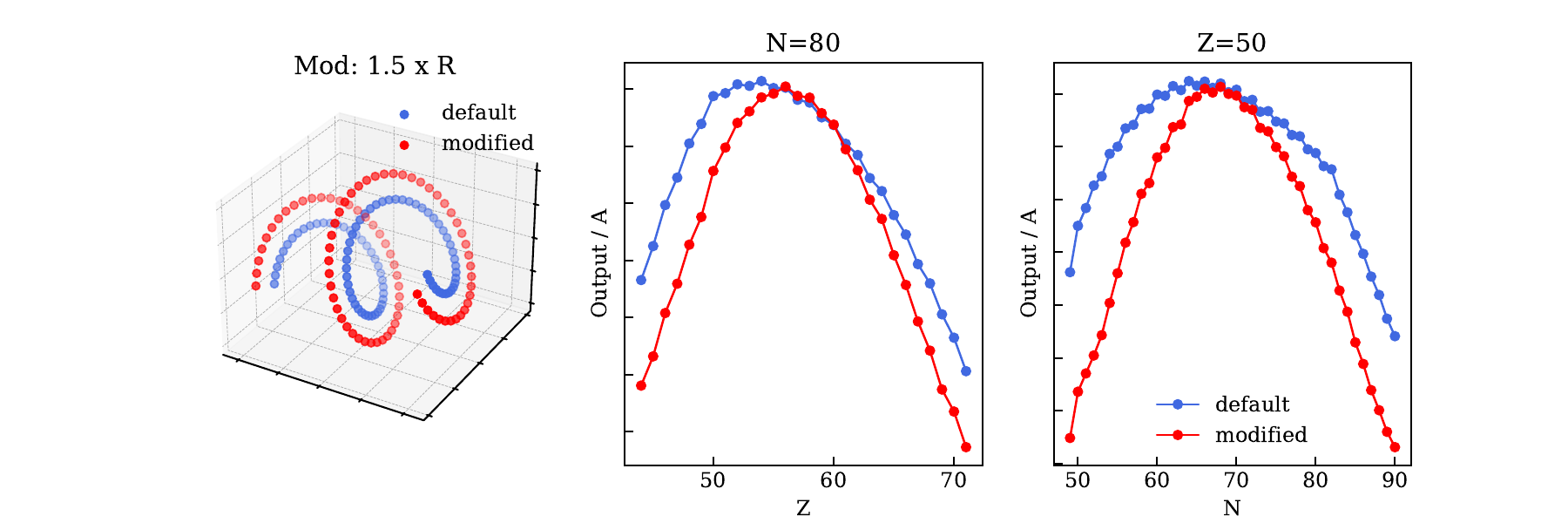}
        \caption{}
    \end{subfigure}
    \hfill
    \begin{subfigure}{.49\linewidth}
        \centering
        \includegraphics[width=\linewidth,trim=100 0 50 0,clip]{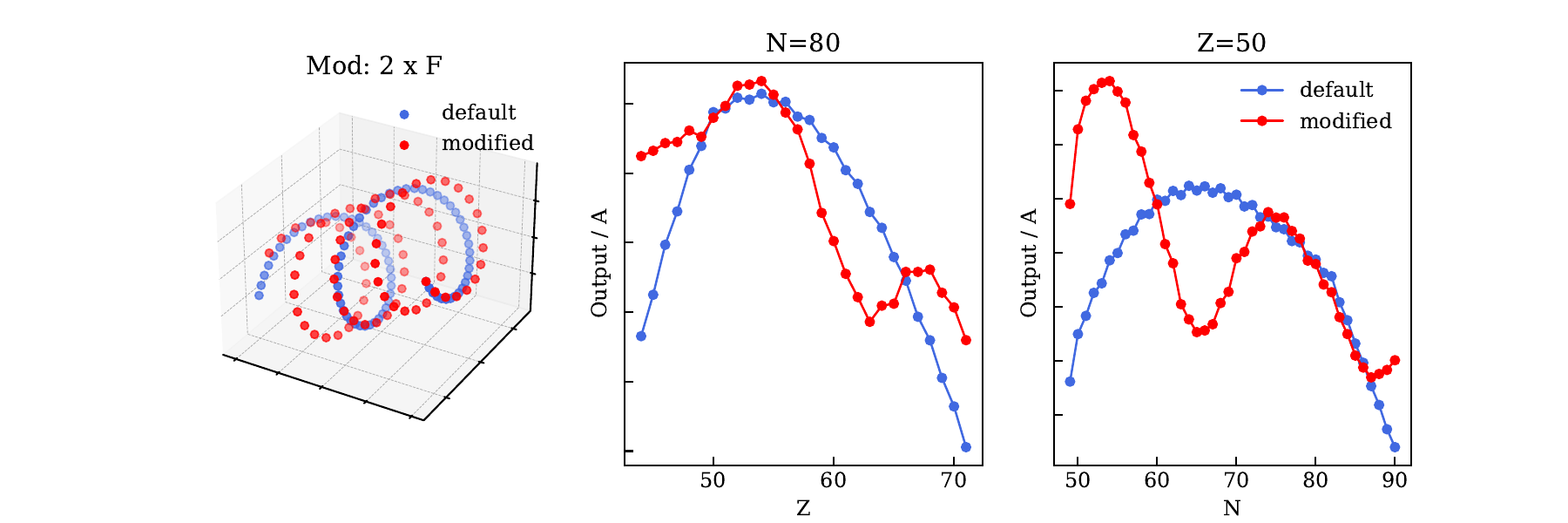}
        \caption{}
    \end{subfigure}
    \begin{subfigure}{.49\linewidth}
        \centering
        \includegraphics[width=\linewidth,trim=100 0 50 0,clip]{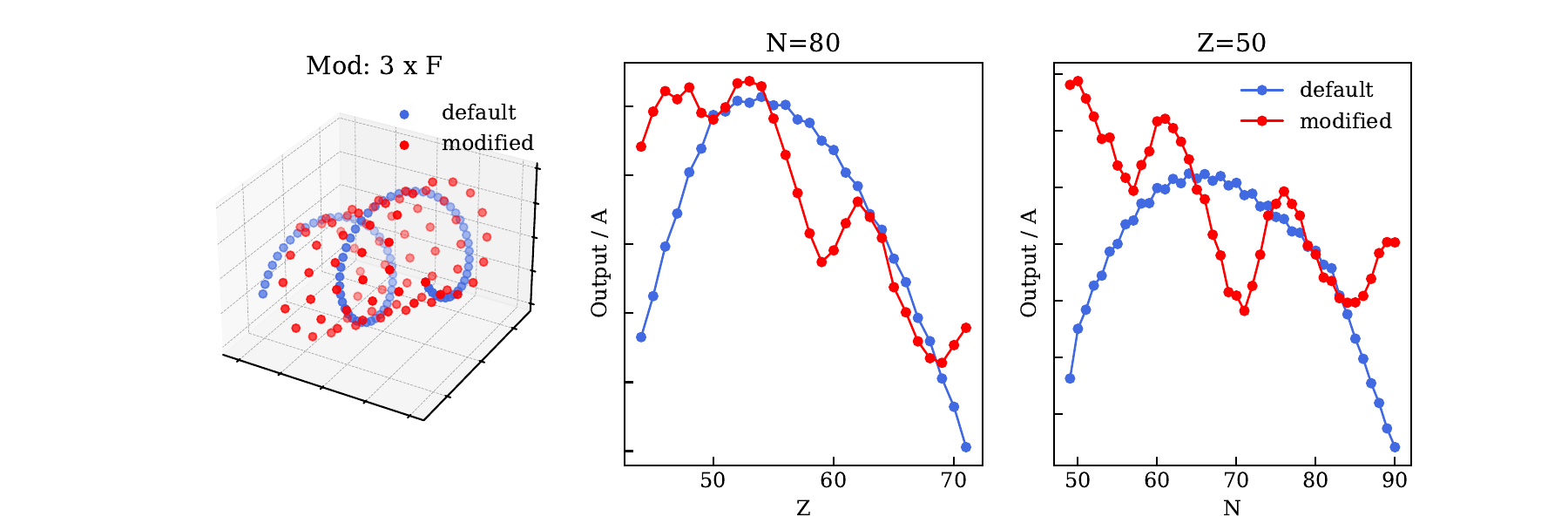}
        \caption{}
    \end{subfigure}
    \caption{Variations in helix parameters and their effects on predictions when: (a) increasing the length by 20\%, (b) reducing the length by 20\%, (c) reducing the radius by 50\%, (d) increasing the radius by 50\%, (e) multiplying the frequency by 2, (f) multiplying the frequency by 3. (Model trained on data).}
    \label{fig:combined_spiral_modifications}
\end{figure}
\clearpage
\vspace{-1em}
\begin{figure}
    \centering
    \includegraphics[width=.49\linewidth,trim=50 50 50 50,clip]{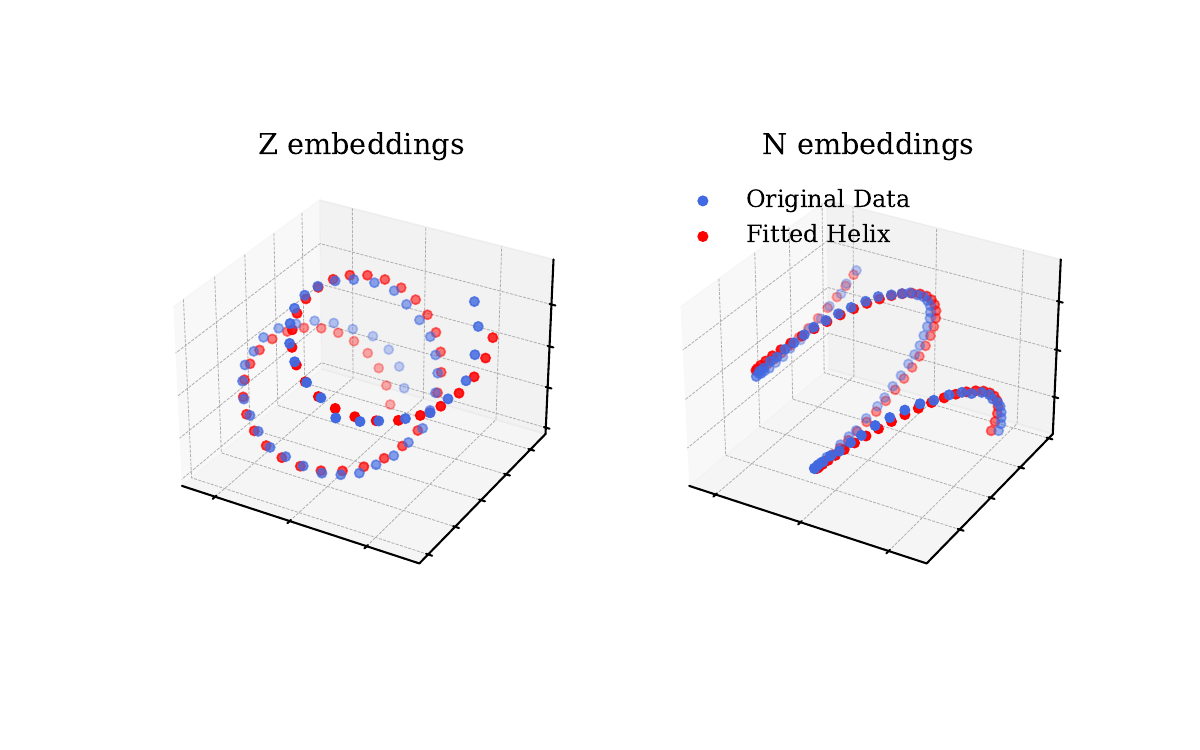}
    \caption{Results of fitting the helix to the selected portions of $N$ and $Z$ embeddings. This model was trained on the SEMF.}
    \label{fig:fit_spirals_semf}
\end{figure}

\begin{figure}
    \centering
    \begin{subfigure}{.49\linewidth}
        \centering
        \includegraphics[width=\linewidth,trim=100 0 50 0,clip]{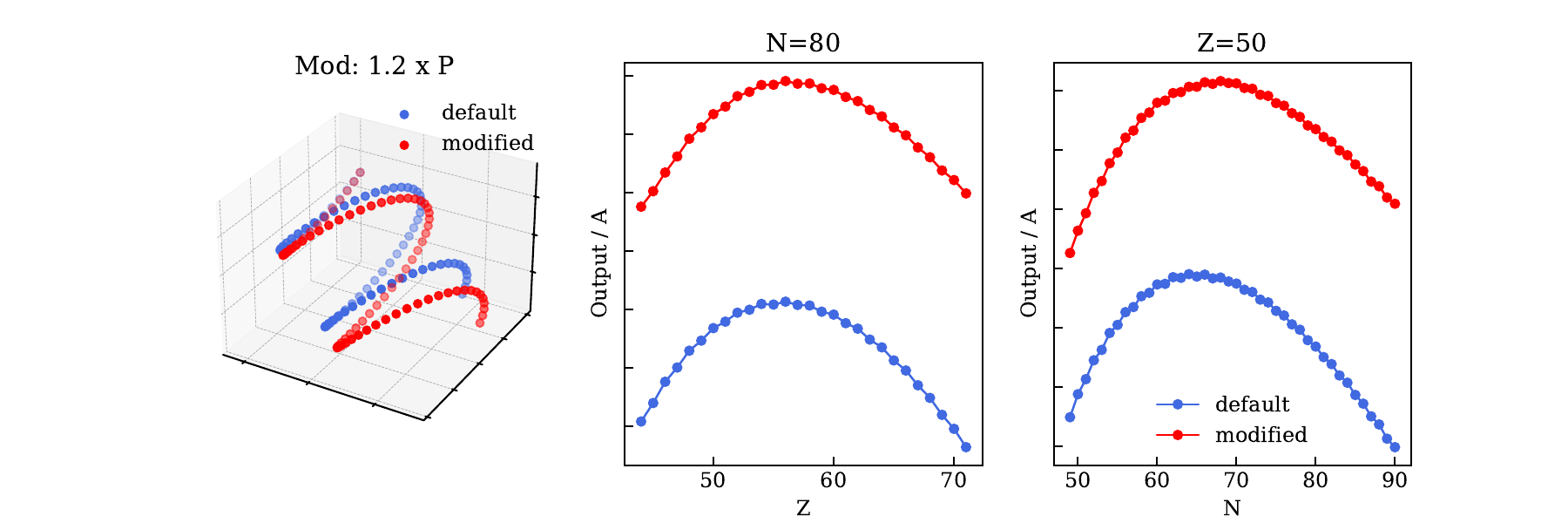}
        \caption{}
        \label{fig:spiral_mod_P_1.2}
    \end{subfigure}
    \begin{subfigure}{.49\linewidth}
        \centering
        \includegraphics[width=\linewidth,trim=100 0 50 0,clip]{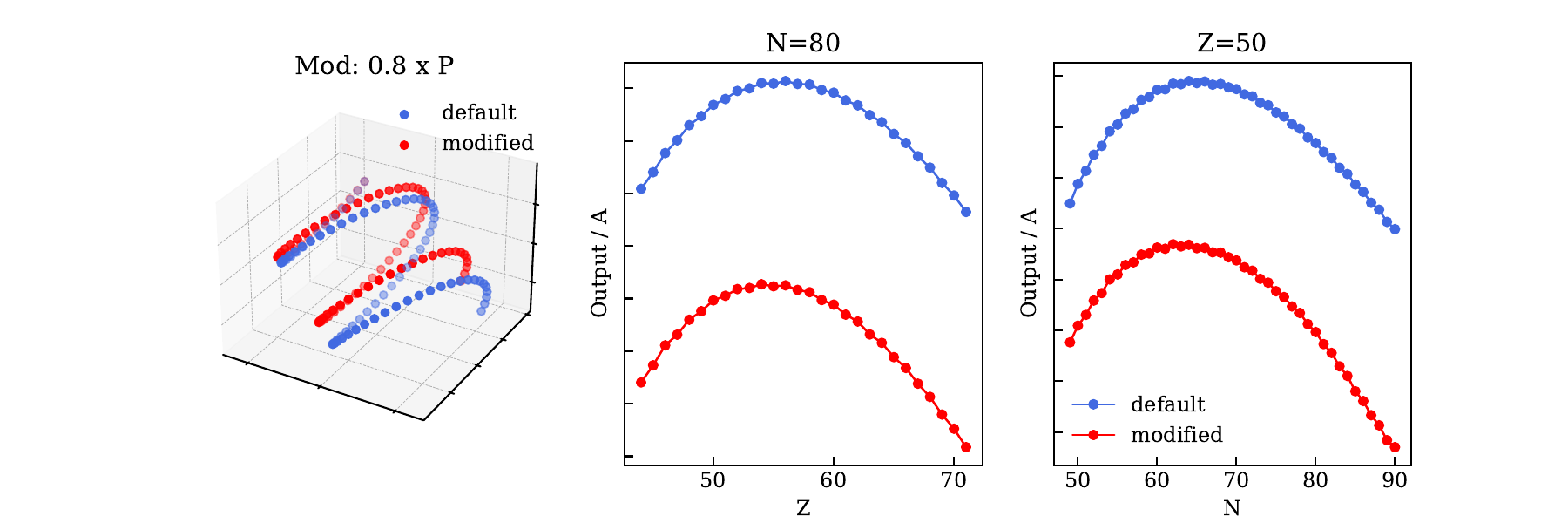}
        \caption{}
        \label{fig:spiral_mod_P_0.8}
    \end{subfigure}
    \hfill
    \begin{subfigure}{.49\linewidth}
        \centering
        \includegraphics[width=\linewidth,trim=100 0 50 0,clip]{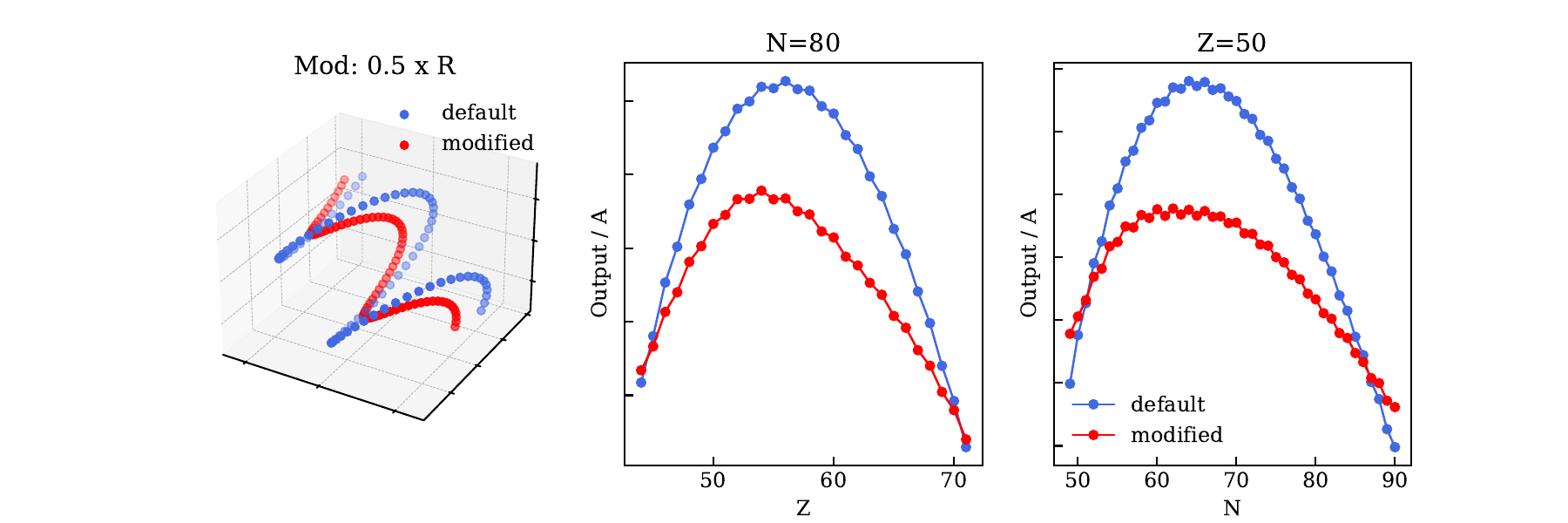}
        \caption{}
        \label{fig:spiral_mod_R_0.5}
    \end{subfigure}
    \begin{subfigure}{.49\linewidth}
        \centering
        \includegraphics[width=\linewidth,trim=100 0 50 0,clip]{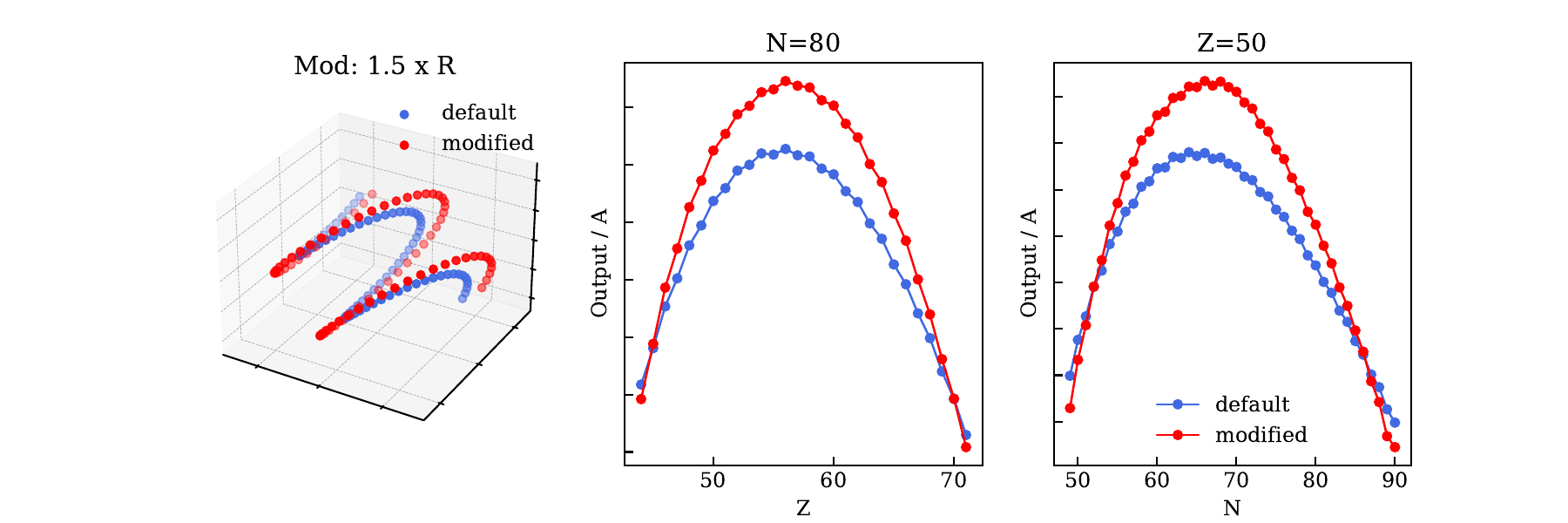}
        \caption{}
        \label{fig:spiral_mod_R_1.5}
    \end{subfigure}
    \hfill
    \begin{subfigure}{.49\linewidth}
        \centering
        \includegraphics[width=\linewidth,trim=100 0 50 0,clip]{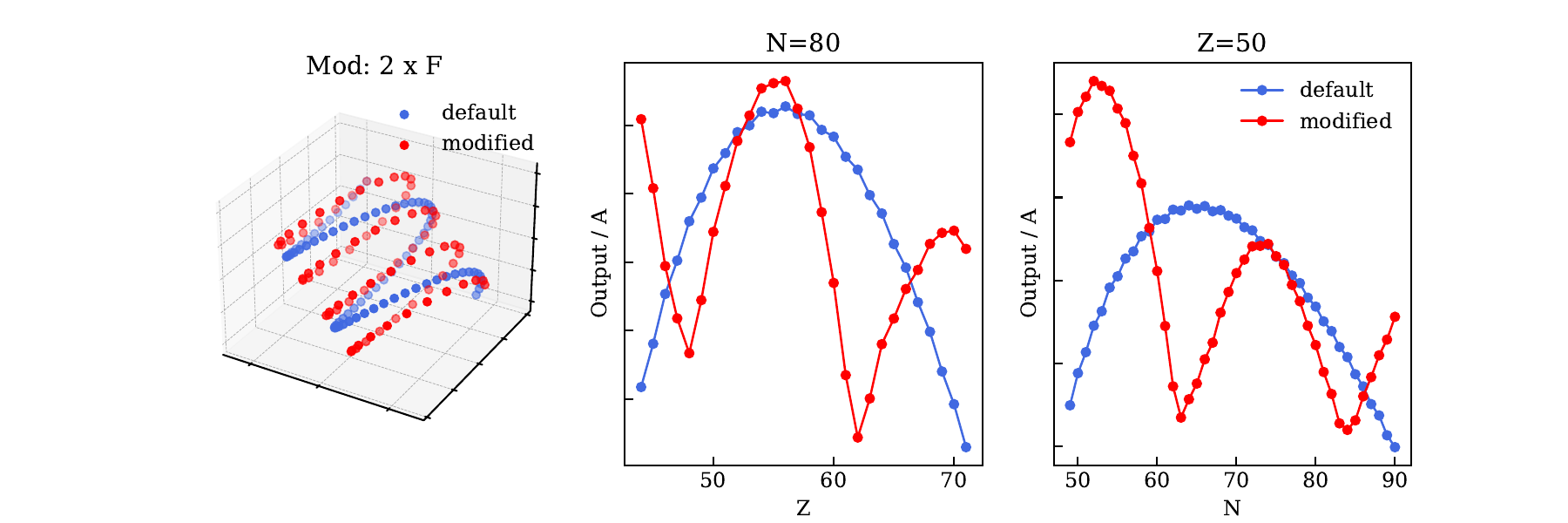}
        \caption{}
        \label{fig:spiral_mod_F_2}
    \end{subfigure}
    \begin{subfigure}{.49\linewidth}
        \centering
        \includegraphics[width=\linewidth,trim=100 0 50 0,clip]{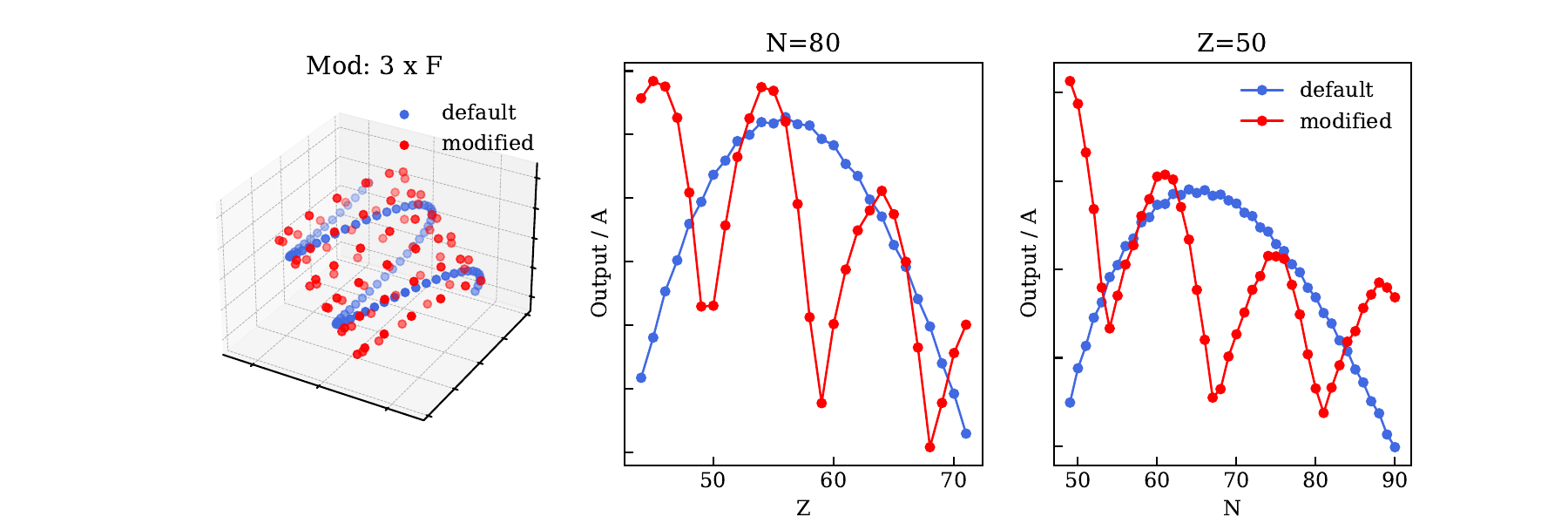}
        \caption{}
        \label{fig:spiral_mod_F_3}
    \end{subfigure}
    \caption{Equivalent of \cref{fig:combined_spiral_modifications}, but for a model trained on the SEMF directly.}
    \label{fig:combined_spiral_modifications_semf}
\end{figure}

\clearpage
\section{Training and model details}
\label{app:training-model-details}
We use an attention ablated transformer with SiLU activations and residual connections. We experimented with different norms (RMS/Layer/Batch)Norm and the results seemed similar to having no norm at all (probably due to shallowness of the models used). Attention seems to matter a lot more despite the fact that model and context length are relatively small. Fixing attention in the way we do can be shown to simplify the model quite drastically \cite{zhong2023clock}. We also found the embeddings to be easier to interpret so we focus on this setup throughout the paper. We use a linear readout layer at the top of the model to predict scalar values which we train with MSE loss. We also experimented with different weighting schemes for the tasks and settled on a ``physics-informed'' scheme based on expected measurement errors for each task.

We use AdamW with mostly default parameters and experiment with a range of hyperparameters in our explorations $\text{learning rate} \in [10^{-4}, 10^{-3}]$, $\text{weight decay} \in [10^{-8}, 10^{-2}]$. 
The runs used to generate the embeddings and visualizations have the following parameters:
\begin{itemize}[nosep]
    \item \texttt{EPOCHS = 200,000}
    \item \texttt{HIDDEN\_DIM = 2048}
    \item \texttt{LR = 0.0001}
    \item \texttt{WD = 0.01}
    \item \texttt{DEPTH = 2}
    \item \texttt{Seed = 0}
\end{itemize}

Most training runs were on Nvidia V100 GPUs with some done on Nvidia A6000 GPUs.
\subsection{Structure evolution}
\label{app:progress-measures}
Here we visualize the progress of our ``strcuture measures" as a function of time for models that generalize well and models that memorize.

\begin{figure}[htbp]
\centering
\begin{subfigure}{.5\linewidth}
    \centering
    \includegraphics[width=\linewidth]{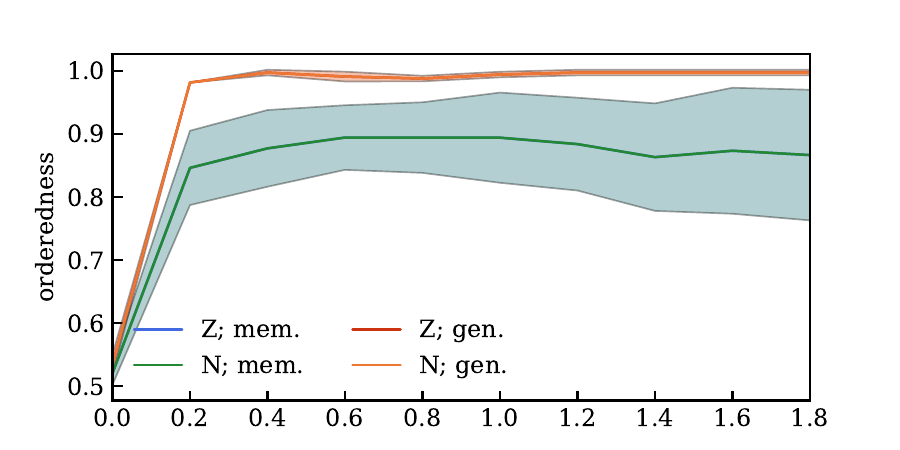}
    \caption{Orderness in time for generalizing and memorizing models.}
    \label{fig:time-order}
\end{subfigure}%
\begin{subfigure}{.5\linewidth}
    \centering
    \includegraphics[width=\linewidth]{plots/figures/parity_split_vs_train_time.pdf}
    \caption{Parity in time for generalizing and memorizing models.}
    \label{fig:time-parity}
\end{subfigure}
\vspace{-1em}
\caption{Progress of structure measures plotted against the number of epochs (normalized by $10^{5}$).}
\label{fig:both}
\end{figure}

\section{Physics models and observables}
\label{app:physics-models}
\subsection{Data}
The data sources are: for the various energies the Atomic Mass Evaluation (AME)~\cite{Wang:2021xhn} and for the charge radii the Atomic Data and Nuclear Data Tables 99 (2013)~\cite{2013ADNDT..99...69A}. We note that all the RMS metrics are calculated using the whole datasets, which include both experimental measurements as well as estimates, e.g. via the method of \textit{trends from the mass surface} (TMS).
\subsection{Liquid-Drop Model (LDM) - the theory behind the SEMF}
\label{app:LDM}

While the properties of the nuclei share the same microscopic origin, namely the strong nuclear force and electromagnetism, experimentally we have access only to a set of macroscopic observables. The first and historically most important nuclear model is the macroscopic LDM, which treats the nucleus as a droplet of highly dense fluid, bound together by the strong nuclear force. The model explains why most nuclei have a spherical shape with a radius proportional to $\sim A^{1/3}$. Impressively, this dependence yields an excellent fit to the charge radius data.

Moreover, the LDM provides an estimation of the binding energy~\cite{semf1935,Bethe:1936zz}, which is the fundamental observable in nuclear physics as it enters the calculations of most of the other quantities. It represents the energy required to break apart a nucleus into its individual nucleons and it is defined as
\begin{equation}
E_B(Z,N) \equiv Z m_p + N m_n - M(Z,N) ~, 
\end{equation}
The LDM prediction for $E_B$ is given by the SEMF (see \eqref{eq:semf}). In the following, we briefly explain the phenomenological motivation for the terms that appear in the SEMF.

\paragraph{Volume Term $+a_V A$:} Represents the bulk energy contribution. The nucleus's overall energy is directly proportional to its volume.

\paragraph{Surface Term $-a_S A^{2/3}$:} Accounts for nucleons on the surface having fewer neighboring nucleons to bond with. It is proportional to the surface area of the nucleus and it is negative, since it corrects the additional contribution assumed for the volume term.

\paragraph{Coulomb Term $-a_C \frac{Z(Z - 1)}{A^{1/3}}$:} Reduces the total energy due the electrostatic repulsion between protons.

\paragraph{Asymmetry Term $- a_S \frac{(N - Z)^2}{A}$} Accounts for the Pauli exclusion principle, i.e. increased energy is required when neutrons and protons are present in unequal numbers, forcing one type of particle into higher energy states.

\paragraph{Pairing Term $\pm a_P A^{-1/2}$}: This term is non-zero only for even $A$ and reflects the stability gained through the pairing of protons and neutrons due to spin coupling. The contribution is either positive or negative if $N$ and $Z$ are both even or odd, respectively.

The SEMF is refined upon the inclusion of a number of additional terms: (i) exchange Coulomb term, (ii) Wigner term, (iii) surface symmetry term, (iv) curvature term, and (v) shell effects term. For detailed explanations of these terms, as well as the fits of all the coefficients $a_\ast$ see \cite{Kirson:2008yvv}. The contributions of these additional terms are depicted in \cref{fig:last-layer-features-2} (the refined SEMF is denoted as BW2).

\subsection{Nuclear shell model}
\label{app:shell_model}

The failure of the SEMF at reproducing the measured values of masses for light nuclei and nuclei with certain numbers of nucleons, the \textit{magic numbers}\footnote{The most widely recognized are $[2, 8, 20, 28, 50, 82, 126]$ and others are still debated.}, led to the development of the \textit{nuclear shell model} by Goeppert-Mayer and Jensen  (Nobel Prize in Physics, 1963). According to this model, protons and neutrons are seperately arranged in shells, and magic numbers occur when shells are filled. Nuclei with either $Z$ or $N$ (or both) equal to a magic (or doubly magic) number exhibit enhanced stability, and thus the $E_B$ spikes. 

The various shell properties can be reproduced by approximating the nuclear potential with a three-dimensional harmonic oscillator plus a spin–orbit interaction. More advanced treatments include the usage of mean field potentials. However, a simple phenomenological term can be still be added to the SEMF and improve its performance. This term is: $a_{M1} P + a_{M2} P^2$, where $P = \frac{\nu_N \nu_Z}{\nu_N + \nu_Z}$ and $\nu_{N,Z}$ the numbers of the valence nucleons (i.e. the difference between the actual nucleon
numbers, $N$ and $Z$ respectively, and the nearest magic numbers). The  contribution of this term can be seen in \cref{fig:matches}.

\subsection{Separation energies}
\label{app:energies}

The stability of a nuclide is determined by its separation energies, which refers to the energies needed to remove a specific number of nucleons from it. They reflect the changes in structure across the nuclear landscape and play a crucial role in understanding the energy requirements involved in nuclear reactions. The separation energies of an isotope can be determined in case the binding energies of neighboring isotopes on the $N-Z$ plane have been measured (and vice-versa). The one-neutron $S_{N}$, one-proton $S_{P}$ separation energy, the energy released in $\alpha$-decay $Q_{\rm A}$, $\beta$-decay $Q_{\rm BM}$, double $\beta$-decay $Q_{\rm BMN}$, and electron-capture process $Q_{\rm EC}$ are, respectively 
\begin{align} \label{eq:sep_energies}
S_N(Z, N) &\equiv M(Z, N - 1) + m_n - M(Z, N)~, \notag \\
S_P(Z, N) &\equiv M(Z - 1, N) + m_p - M(Z, N)~. \notag \\
Q_{\rm A}(Z, N) &\equiv M(Z, N) - M(Z - 1, N + 1) - m_{^4_2 \rm He}~ \notag \\
Q_{\rm BM}(Z, N) &\equiv M(Z, N) - M(Z + 1, N - 1)~,  \notag \\
Q_{\rm BMN}(Z, N) &\equiv M(Z, N) - m_n - M(Z + 1, N - 2)~,  \notag \\
Q_{\rm EC}(Z, N) &\equiv M(Z, N) - M(Z - 1, N + 1)~.
\end{align}

\begin{figure}[htbp]
    \centering
    \includegraphics[width=0.5\linewidth,trim=0 0 0 0,clip]{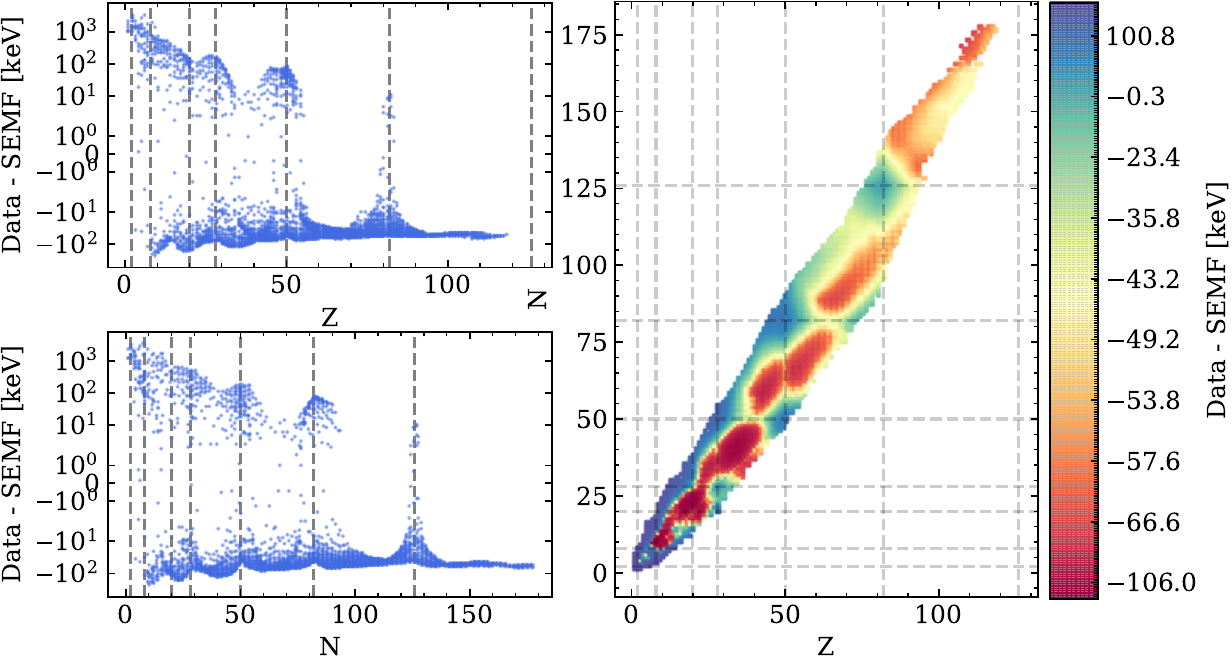}
    \caption{Residual between data and the semi-empirical mass formula. Dashed lines are magic numbers.}
    \label{fig:semf-residual}
\end{figure}

\section{Which representations come from which task?}
\label{app:which-task-rep}
\begin{figure}[htbp]
    \centering
    \includegraphics[width=0.5\linewidth]{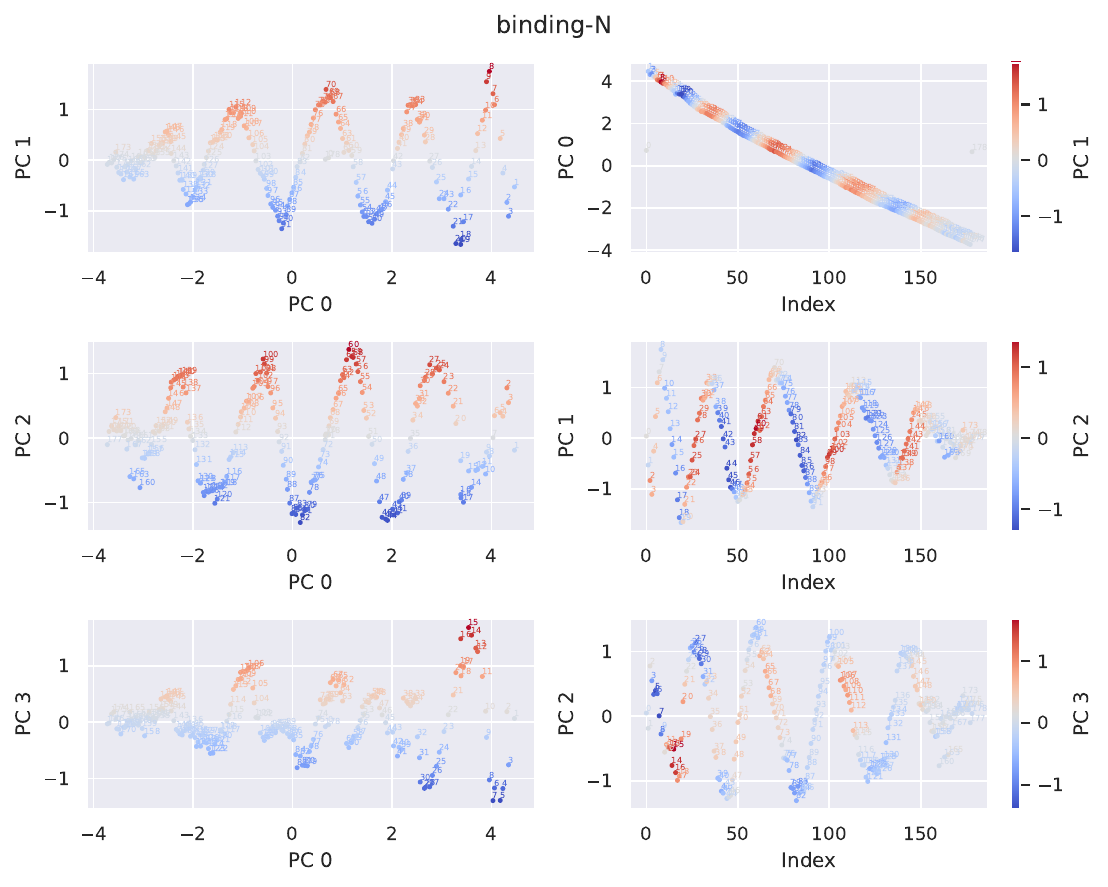}
    \caption{First few PC projections of the $N$ embeddings for a model trained on only binding energy. Index here refers to the token index or the value of $N$.}
\end{figure}

\begin{figure}[htbp]
    \centering
    \includegraphics[width=0.5\linewidth]{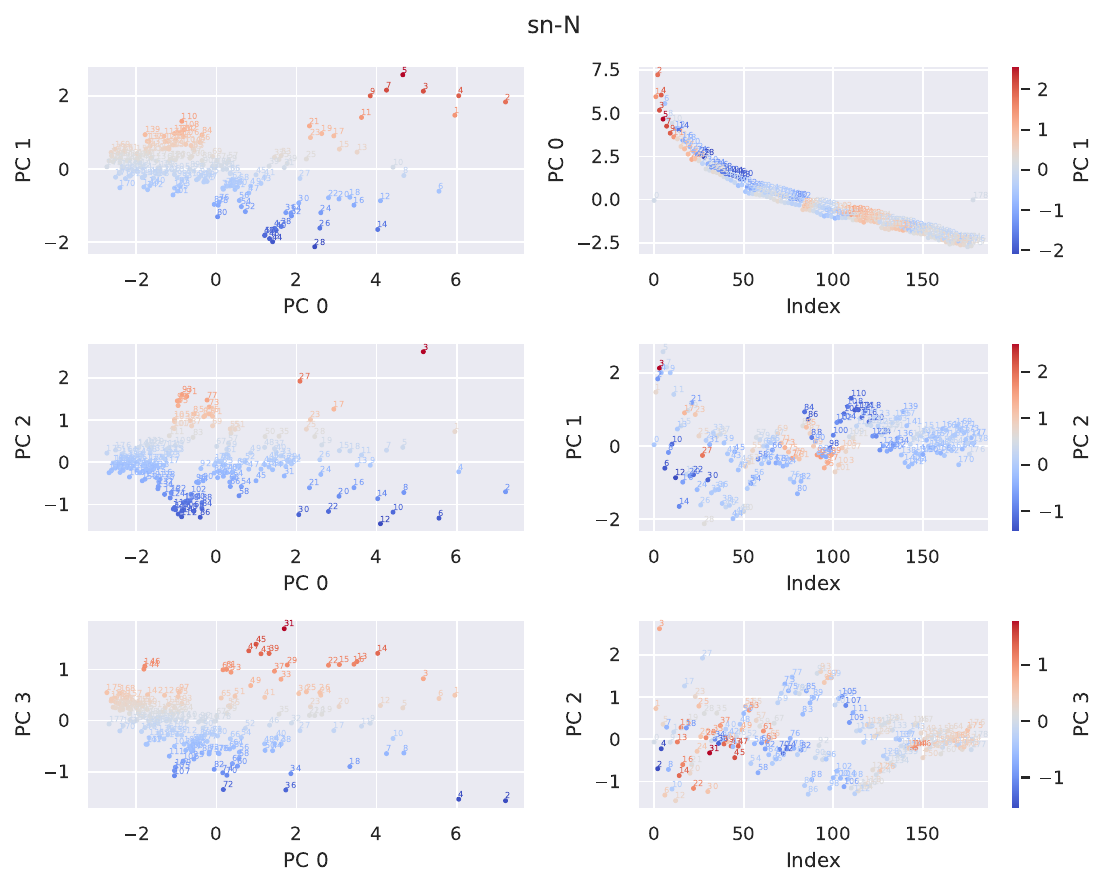}
    \caption{First few PC projections of the $N$ embeddings for a model trained on the target $S_N$ only.}
\end{figure}

\begin{figure}[htbp]
    \centering
    \includegraphics[width=0.5\linewidth]{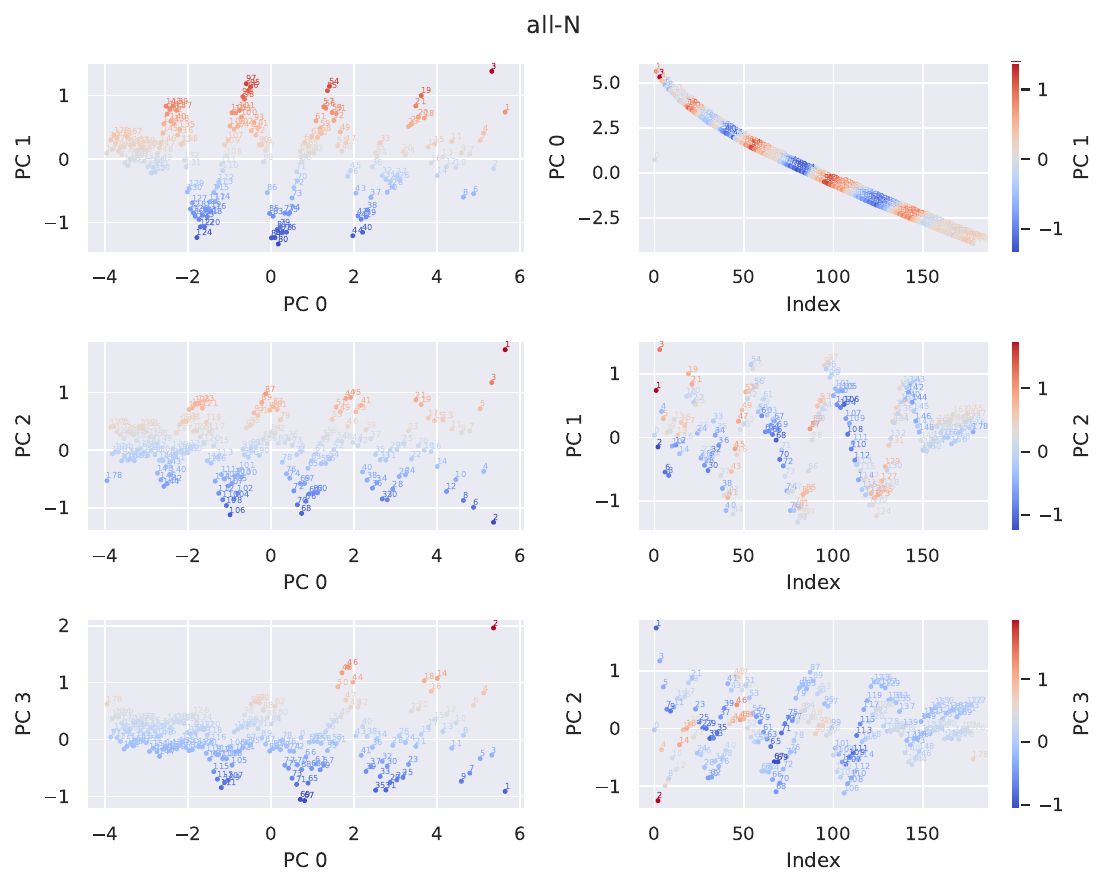}
    \caption{First few PC projections of the $N$ embeddings for a model trained on ``all" data \ie, in the multi-task setting.}
\end{figure}

\clearpage
\section{Penultimate layer features}
\begin{figure}[htbp]
    \centering
    \includegraphics[width=\linewidth]{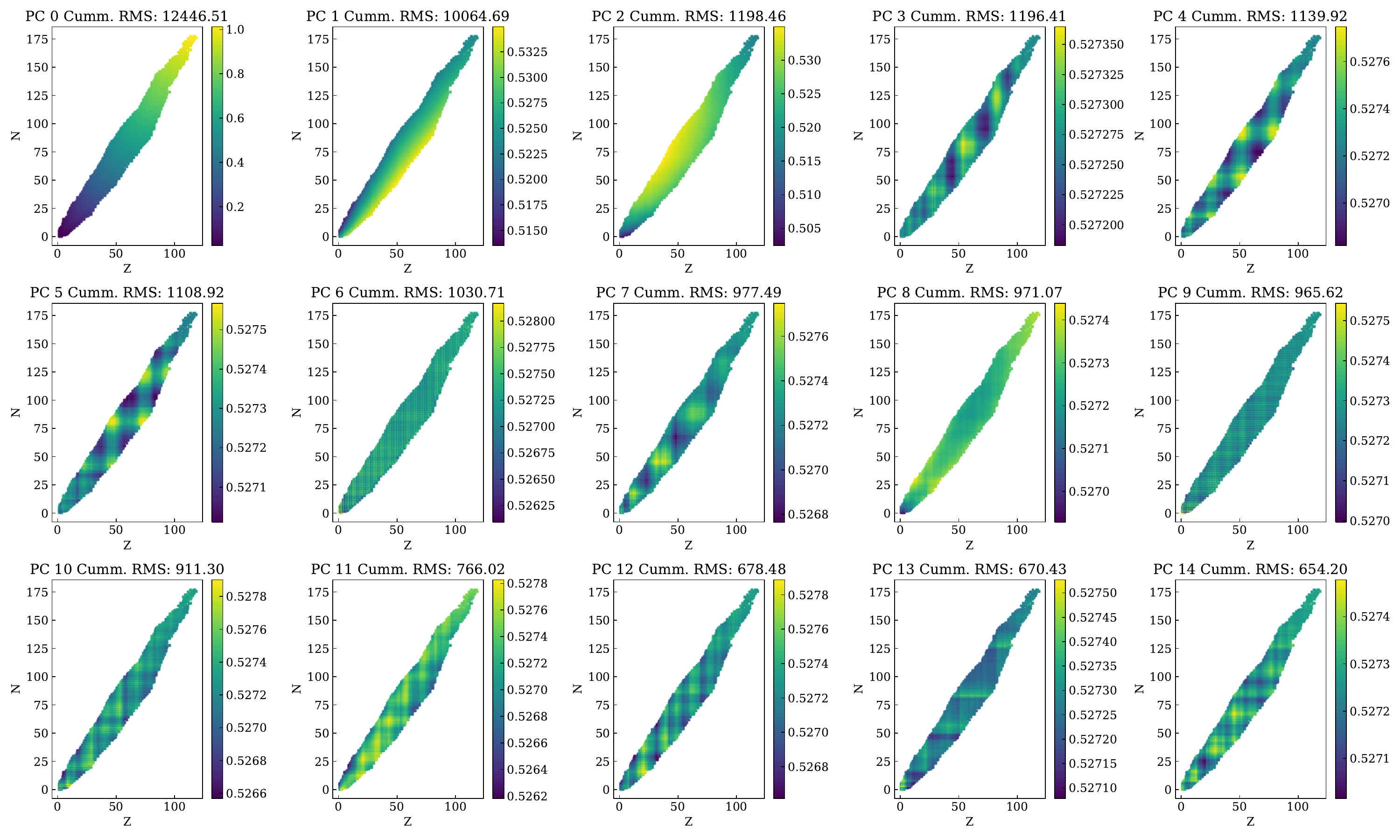}
    \caption{Visualization of of a few penultimate layer PC features and their cumulative effect on the error in binding energy prediction (the error is computed up to and including the PC).}
    \label{fig:last-layer-features}
\end{figure}
\begin{figure}[htbp]
    \centering
    \includegraphics[width=\linewidth]{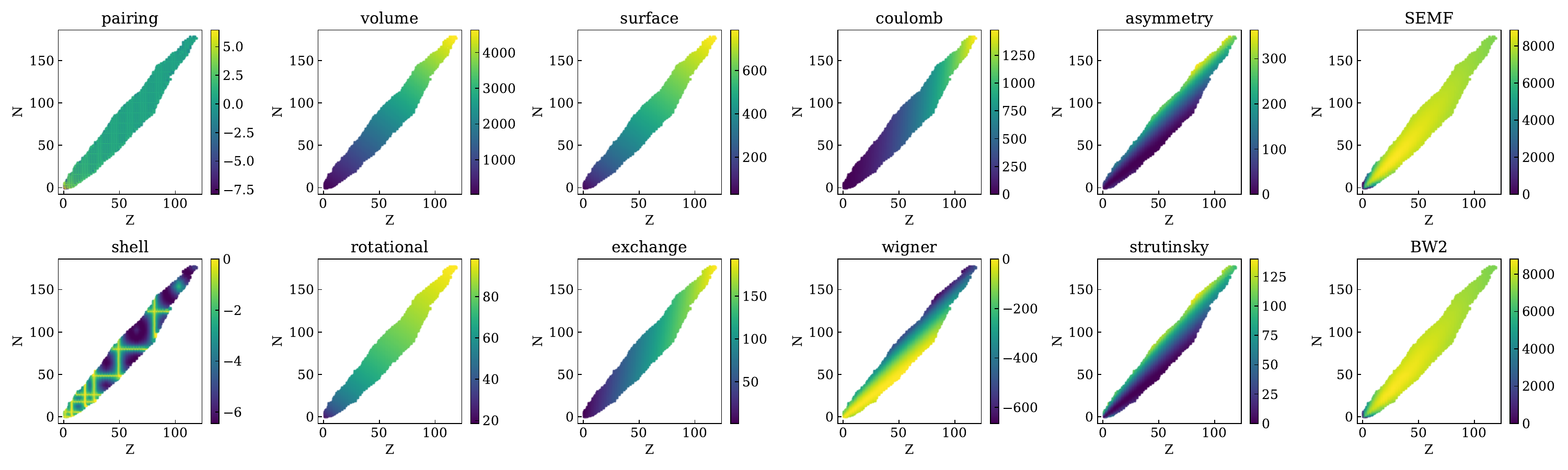}
    \caption{Physics terms visualized. The top row are the terms from the SEMF. The bottom row includes nuclear shell model corrections (BW2 terms).}
    \label{fig:last-layer-features-2}
\end{figure}
\begin{figure*}
    \centering
    \includegraphics[width=0.8\linewidth,]{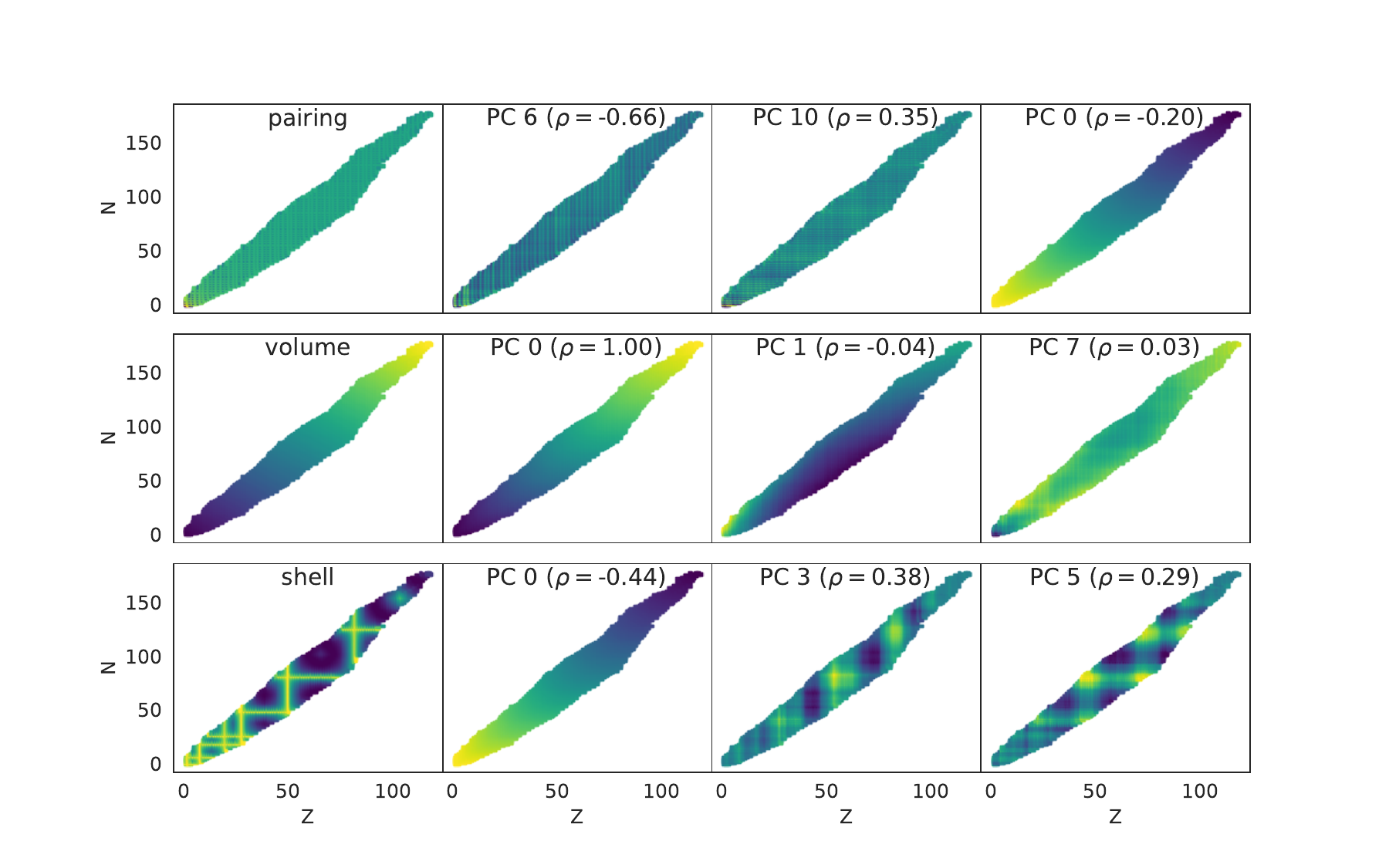} 
    \vspace{-1em}
    \caption{Model penultimate features in the multi-task setting. Physical terms derived from the Nuclear Shell Model and their best matching PCs.}
    \label{fig:matches}
\end{figure*}

\clearpage
\section{Other structures}
\label{app:linear-continuity}
We discussed how the helix structure (essentially stacked circles) is ideal to model the continuous spectrum of binding energies.
However, continuity can be realized in other ways than in a circle (or helix when considering PC0), for instance by a simple line. In fact, we believe that the circular structure is chosen by the model because weight decay favors a continuous structure if it revolves around 0. A circular structure presents a good trade off between embedding weight norm and sufficient distance between elements to form separate predictions for each $Z$ or $N$ without resorting to high weight norm in other layers. \cref{fig:no_wd_embs} shows $N$ embedding projections from a model trained without weight decay, but with somewhat comparable test set performance. As hypothesized, a continuous structure emerges, but no helix. This behaviour is conceptually consistent over different random seeds.

\begin{figure}[htbp]
    \centering
    \includegraphics[width=\linewidth,trim=0 25 0 40,clip]{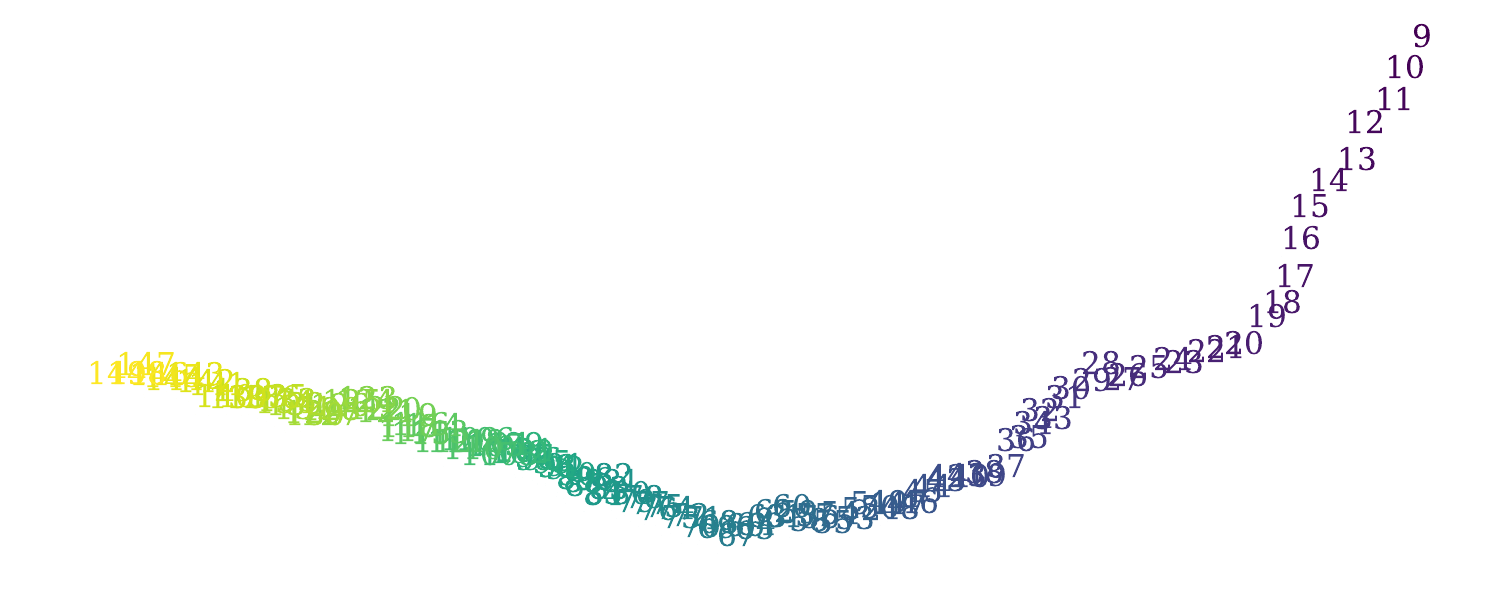}
    \caption{Neutron embeddings projected into the first two PC from a model trained without weight decay.}
    \label{fig:no_wd_embs}
\end{figure}

\section{Symbolic regression}
\label{app:symbolic}

We use symbolic regression to find functions $f_{\text{PC}}^i(Z,N)$ that map from $Z$ and $N$ to the $i$-th feature extracted from the penultimate layer. We use the \texttt{PySR} library \citep{cranmer2023interpretable}, which employs an evolutionary tree-based algorithm.\footnote{In the physical sciences, this method has proven useful for extracting symbolic formulas that reveal new physical patterns or reinterpret known physical laws \cite{mengel2023interpretable,davis2023discovery,lemos2023rediscovering}.}, 

Subsequently, we may write the new expression for the binding energy as $E_B = \sum_{i=1}^{n_F} a_i f_{\text{PC}}^i(Z,N) + b$, where $n_F$ is the number of PC features that are used. The coefficients $a_i$ and the intercept $b$ are determined using linear regression on the binding energy dataset without the TMS values. We find that the using the fits of solely PC0 and PC2, we can retain the bulk of the prediction. The new expression for binding energy reads,

\begin{equation}
  E_B = a_1 \left(-0.09 + 10^{-6} Z^2\right) \left[A + 2.5 \sin\left(0.25 - 0.13N + 0.2Z\right)\right] + a_20.97^N + b~.  
\end{equation}

where $a_1 = -88062.52$, $a_2 = -171331.53$ and $b = 95815.44$.
This formula achieves an RMS of around $4600$ keV. As a comparison, the performance of the SEMF over the same dataset is $8000$ keV. Noteably, any direct regression on the data leads to considerably worse predictions for the same number of free parameters. We assess thus, that the analysis of the representation space of neural networks may streamline symbolic regression tasks.

\section{Limitations}
The interpretability of the extracted knowledge is not guaranteed. Even if the network finds a low-rank structure, it may not necessarily correspond to a simple, interpretable theory that provides clear insight to domain experts. The learned representations might capture complex, nonlinear interactions that are hard to distill into compact, explainable expressions. Moreover, there is currently a lack of quantitative metrics to assess the interpretability of the extracted knowledge. Developing such metrics is crucial, as that which is measured can be improved. Without a way to quantify interpretability, it becomes challenging to track progress and iterate on techniques to enhance the clarity and usefulness of the derived insights for domain experts. As seen in the attempts at symbolic regression, the expressions recovered from the neural features did not yield fully interpretable improvements over human-derived models. This limitation highlights the need for more rigorous metrics to guide the search for more explainable and meaningful representations of the learned knowledge.

Additionally, integrating MI into the scientific discovery workflow requires interdisciplinary collaborations and close partnerships between machine learning researchers and domain experts. Translating between the language of neural network components and the scientific concepts of a given field is a significant challenge that demands dedicated effort from both sides to have a real-world impact in driving scientific progress.

\end{document}